\pdfoutput=1
\documentclass[11pt]{article}

\usepackage[margin=1in]{geometry}
\usepackage{graphicx}
\usepackage{amsmath,amssymb}
\usepackage{booktabs}
\usepackage{array}
\usepackage{tabularx}
\usepackage{longtable}
\usepackage{ltablex}
\usepackage{multirow}
\usepackage{makecell}
\usepackage{nicefrac}
\usepackage{ragged2e}
\usepackage{fvextra}
\usepackage{hyperref}
\usepackage{cleveref}
\usepackage[numbers,sort&compress]{natbib}
\usepackage[table]{xcolor}
\usepackage{microtype}
\usepackage{caption}

\usepackage{amsfonts}       
\keepXColumns

\newcolumntype{Y}{>{\RaggedRight\arraybackslash\hspace{0pt}}X}

\DefineVerbatimEnvironment{verbatim}{Verbatim}{breaklines=true}

\graphicspath{ {./images/} }

\title{Combining LLMs and Knowledge Graphs to Reduce Hallucinations in Biomedical Question Answering}

\author{
  Larissa Pusch\\
  Zuse Institute Berlin, Berlin, Germany\\
  \texttt{pusch@zib.de}
  \and
  Tim OF Conrad\\
  Zuse Institute Berlin, Berlin, Germany\\
  \texttt{conrad@zib.de}
}
\date{} 

\begin{document}
\maketitle
\begin{abstract}
    Advancements in natural language processing (NLP), particularly Large Language Models (LLMs), have greatly improved how we access knowledge. However, in critical domains like biomedicine, challenges like hallucinations - where language models generate information not grounded in data - can lead to dangerous misinformation. This paper presents a hybrid approach that combines LLMs with Knowledge Graphs (KGs) to improve the accuracy and reliability of question-answering systems in the biomedical field. Our method, implemented using the LangChain framework, includes a query-checking algorithm that checks and, where possible, corrects LLM-generated Cypher queries, which are then executed on the Knowledge Graph, grounding answers in the KG and reducing hallucinations in the evaluated cases. We evaluated several LLMs, including several GPT models and Llama 3.3:70b, on a custom benchmark dataset of 50 biomedical questions. GPT-4 Turbo achieved 90\% query accuracy, outperforming most other models. We also evaluated prompt engineering, but found little statistically significant improvements compared to the standard prompt, except for Llama 3:70b improving with few-shot prompting. To enhance usability, we developed a web-based interface that allows users to input natural language queries, view generated and corrected Cypher queries, and inspect results for accuracy. This framework improves reliability and accessibility by accepting natural language questions and returning verifiable answers directly from the knowledge graph, enabling inspection and reproducibility.
The source code for generating the results of this paper and for the user-interface can be found in our Git repository: https://git.zib.de/lpusch/cyphergenkg-gui
\end{abstract}

\section{Introduction}

Advances in natural language processing (NLP) have improved access to knowledge, allowing users to retrieve it in everyday language. In high-stakes areas like biomedicine, however, these advancements introduce new challenges, especially the risk of hallucinations \cite{xu2024hallucination,Sequeda2025,huang2025survey,Wang2024}, where models generate information unsupported by the underlying data. Inaccuracies in biomedical contexts can lead to harmful misinformation, such as incorrect treatments or diagnoses.

Biomedical databases, often organized as Knowledge Graphs (KGs), model complex data like diseases, drugs, and proteins but remain challenging for non-experts to query due to specialized query languages. This paper addresses the challenge of reliable natural language question answering over biomedical KGs, focusing on reducing hallucinations. We propose a novel approach that integrates Large Language Models (LLMs) with KGs, enhancing the accuracy and reliability of responses by grounding answers in structured data.

Our method builds on the LangChain \cite{langchain} framework to convert natural language into Cypher, the query language for graph databases like Neo4j. An important feature is a query-checking algorithm that validates LLM-generated Cypher queries to ensure they align with the KG’s schema, reducing errors and improving reliability. Validated queries are executed on the KG, and the results are used to generate a natural language response via retrieval-augmented generation (RAG) \cite{rag}.

As an example, we applied our method to PrimeKG \cite{chandak2022building}, a biomedical KG encompassing data on diseases, drugs, and proteins. Our system allows users to pose questions, view Cypher queries, and examine results through a user-friendly web interface. We evaluated this approach by doing a systematic comparison of LLMs and prompt strategies with a custom benchmark dataset of 50 biomedical questions, testing several LLMs, including GPT-4 Turbo, GPT-5, and Llama 3.3:70b. Results show that GPT-4 Turbo yields the most accurate queries, while the 70b LLama3 models lead the field of open source models tested in this benchmark.

In summary, our pipeline combines LLMs and KGs to address challenges such as hallucinations, providing accurate biomedical question answering and making advanced query capabilities accessible to non-experts.

\subsection{Problem Statement and Research Goals}

This paper addresses key challenges in using Large Language Models (LLMs) for accurate question answering over Knowledge Graphs, specifically tackling data gaps and hallucinations, where models produce ungrounded or incorrect information.

Our research focuses on these main objectives: \begin{itemize}

\item \textbf{Reducing Data Gaps and Hallucinations}: We aim to decrease inaccuracies and fabrications in LLM responses by integrating them with Knowledge Graphs and using a query-checking algorithm that verifies and corrects Cypher queries generated by LLMs, targeting common syntactic and schema alignment issues. Additionally, we optimize prompts to guide LLMs in producing more accurate queries, enhancing response reliability.

\item \textbf{Evaluating LLM Performance}: We assess the performance of several LLMs, including GPT-4 Turbo and Llama 3:70b, on a custom benchmark dataset to identify model strengths and weaknesses and explore improvements for open-source models via prompt engineering.

\item \textbf{Creating a Benchmark Dataset}: We developed a dataset of 50 biomedical questions based on a subset of PrimeKG for evaluating LLMs’ ability to generate accurate Cypher queries for a specific biomedical Knowledge Graph, providing a foundation for future research in this domain.

\item \textbf{Designing a User-Friendly Interface}: To make the system accessible, we created a user-friendly web-based interface where users can input natural language queries, view generated and corrected Cypher queries, and inspect results.

\end{itemize} Our overarching goal is to combine the accessibility of LLMs and the reliability of complex digital information systems, specifically, Knowledge Graphs, to enable non-expert users to benefit from them.

\section{A New Approach for LLM-based Knowledge Graph Queries}

\label{pipeline_details}

In this section, we introduce our approach designed to address the challenges associated with using Large Language Models (LLMs) for information retrieval. 

The overall method consists of three main steps: 
\begin{enumerate}
\item The user's question, along with the graph schema, is passed to the LLM, which generates a Cypher query. 
\item This generated query is then subjected to the query-checking algorithm for validation and potential correction. 
\item Finally, the validated query is executed on the Knowledge Graph, and the results are returned.
\end{enumerate}

\subsection{Step 1: Initial Cypher Query Generation}
\label{cypher_query_gen}
The initial Cypher query is generated by the LLM using a prompt that includes the user query and the full graph schema, which is created with LangChain Neo4jGraph \cite{langchain_graph}. We use exact matching by design to isolate reasoning-to-Cypher under a fixed schema. Adding semantic or fuzzy matching would change the construct to alias coverage and is out of scope for this evaluation. Consistent with this design, our study is a systematic comparison of LLMs and prompt strategies for KG querying under a fixed schema, not a global leaderboard across heterogeneous KGQA paradigms. All models are evaluated identically on the same KG snapshot and parsing pipeline. The LLM outputs a Cypher query to retrieve answers from the Knowledge Graph, choosing node and relationship types from the schema. More information about the prompts can be found in \autoref{llm_comparison}. If the message consisted of a single fenced Cypher code block, we used its contents after removing the fence. Otherwise we extracted the first contiguous span starting at the first line containing MATCH and ending at the first line containing RETURN, discarding any text before or after that span. This procedure was applied identically across models. Outputs without a Cypher block or a MATCH ... RETURN span were marked invalid. 

\subsection{Step 2: Query Checker}
\label{query_checker}
The query checker checks common schema and direction issues before execution and can correct near-miss errors; it is designed to improve validity, not to serve as a full verifier. This process involves three components, each targeting specific aspects of query validation.

Consider the example query: "What are the names of the drugs that are contraindicated when a patient has multiple sclerosis?" A typical LLM-generated Cypher query might look like this:

\begin{verbatim}
MATCH (d:pathway {name:"multiple sclerosis"})-[:contraindication]->(dr:drug)
RETURN dr;
\end{verbatim}

However, this translation has the following issues:

\begin{enumerate}
     
\item "Name" attribute missing from the returned node: The output \textbf{RETURN dr;} would return the entire drug node, including all its properties, rather than just the drug names. Thus, the syntax checker would refine the query to \textbf{RETURN dr.name;}. 
\item Wrong node type: \textbf{(d:pathway \{name:"multiple sclerosis"\})} The LLM incorrectly identifies "multiple sclerosis" as a pathway, instead of a disease. The node checker would correct this error by modifying the Cypher query to: \textbf{(d:disease \{name:"multiple sclerosis"\})}
\item Relationship direction error: The query incorrectly directs the "contraindication" relationship as \textbf{-[:contraindication]->}, pointing from the disease to the drug. The correct direction should have the relationship pointing from the drug to the disease. This will be corrected by the relation checker to: \textbf{<-[:contraindication]-}.
\end{enumerate}

In summary, the query checker executes the following three steps:

\begin{itemize} 
\item Syntax Node Checker: This component ensures that the output of the Cypher query returns the resulting node names by appending the \textit{.name} property to each node in the return statement. It also verifies that any variables specified in the return clause are correctly associated with their respective node types in the MATCH statement, in the form of \textit{node: node\_type}. 
\item Node Checker: This checker validates the types of nodes referenced in the query, ensuring they are the correct type for the item that was extracted from the question. If an incorrect node type is identified, it automatically substitutes the correct type. It also evaluates whether the relationships involving the corrected node type are appropriate and adjusts them if necessary, unless no compatible relationships exist, in which case the step is skipped.
\item Relation Checker: This component verifies the directionality of relationships between nodes, ensuring that they are oriented correctly within the query. If any relationships are found to be reversed, the Relation Checker automatically corrects their direction to maintain the integrity of the query's logic.
\end{itemize}

\subsection{Step 3: Querying the Knowledge Graph}
The Cypher query, generated by the LLM and verified by the query checker, is executed on the Knowledge Graph using Neo4jGraph from LangChain. Then, an answer sentence can be generated by the LLM based on the returned list. This retrieval-augmented generation (RAG) grounds each answer in the Knowledge Graph.

\section{Data}
\label{data}
This section describes the data used in our experiments, including the Knowledge Graph and a custom set of questions and answers.

\subsection{PrimeKG-based Knowledge Graph}

We used a subset of the biomedical Knowledge Graph PrimeKG \cite{chandak2022building}, specifically a 2-hop subgraph around multiple sclerosis. This subset contains 44,348 triples, 22 unique relations, and 7,413 entities, covering node types such as drug, disease, phenotype, gene/protein, anatomy, cellular component, pathway, molecular function, exposure, and biological process. To optimize query execution, we made all relations unidirectional (except self-connections). Further details can be found in \autoref{primekg_changes}.

\subsection{Tailored Question/Answer Set}
\label{questions}

Traditional question datasets often don’t adequately assess a pipeline's performance in querying Knowledge Graphs, but rather the graph completeness, as they are not tailored to the graph in question. To address this, we developed a custom set of questions specifically tailored for the selected Knowledge Graph. These questions are directly answerable using the graph’s data (see \autoref{question_appendix} for the original set and \autoref{sec:paraphrased} for the paraphrased set). Initially, we identified paths that met our structural criteria and included only those with non-empty result sets (to make it possible to automatically assess correctness based on the answers).
Questions were created using 1, 2, and 3-hop paths, representing five distinct structures. In Knowledge Graphs, a 'hop' refers to the number of edges between nodes; a 1-hop represents a direct connection, while higher hops indicate more complex queries involving intermediate nodes and relationships. These different structures allowed us to test the pipeline's ability to handle queries of varying complexity.
The dataset is designed to test schema comprehension and query construction over 1-, 2-, and 3-hop patterns in PrimeKG, not recall of disease facts or entity-specific knowledge.

During testing, we found that LLMs sometimes identified alternate, valid paths for answering questions that weren’t initially considered. These alternative answers, validated through manual curation, showcased the models' flexibility in navigating the Knowledge Graph.

The way we structured questions into five types based on 1, 2, or 3-hop paths is outlined below.

\subsubsection{1-hop}
\begin{figure}
  \centering
  \includegraphics[scale=0.7]{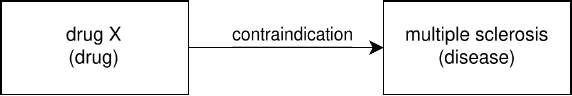}
  \caption{Example structure for 1-hop question}
  \label{fig:q1}
\end{figure}

For 1-hop questions, there is a straightforward structure involving a direct relationship between two entities. This allows for simple, direct queries, see \autoref{fig:q1}.
\begin{itemize}
\item Structure 1: A single, direct relationship between two nodes, typically excluding bidirectional relations. Example question: \textit{What are the names of the drugs that are contraindicated when a patient has multiple sclerosis?}
\end{itemize}

\subsubsection{2-hop}
2-hop questions allow for slightly more complex queries, involving either a single entity directly related to two others or a linear arrangement (chain) of three entities.

\begin{itemize} 
\item Structure 2: One entity is directly connected to two other entities, see \autoref{fig:q2_s1}. The question for this path was \textit{What side effects does a drug have that is indicated for Richter syndrome?}
\begin{figure}
  \centering
  \includegraphics[scale=0.7]{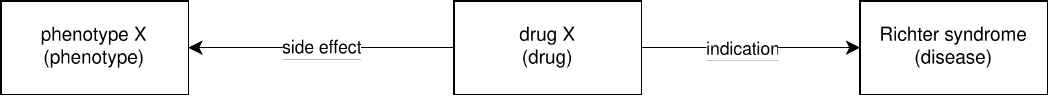}
  \caption{Example structure for 2-hop question (structure 2)}
  \label{fig:q2_s1}
\end{figure}
\item Structure 3: A linear arrangement (chain) where one entity is connected to a second, which in turn is connected to a third, see \autoref{fig:q2_s2}. The question for this path was \textit{What are phenotypes that gene POMC is associated with that also occur in neuromyelitis optica?}
\begin{figure}
  \centering
  \includegraphics[scale=0.7]{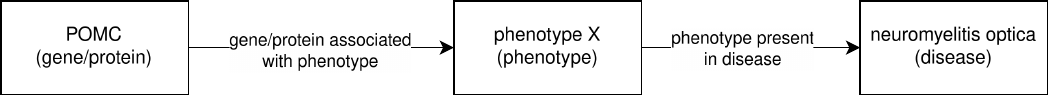}
  \caption{Example structure for 2-hop question (structure 3)}
  \label{fig:q2_s2}
\end{figure}
\end{itemize}

\subsubsection{3-hop}
3-hop questions introduce the highest complexity, involving either a sequential chain of four items or a configuration where an entity interfaces with multiple others in a more extended arrangement.

\begin{itemize} 
\item Structure 4: A sequential chain of four connected items, see \autoref{fig:q3_s1}. The question for this path was \textit{What pathways do the exposures that can lead to multiple sclerosis interact with?}
\begin{figure}
  \centering
  \includegraphics[scale=0.7]{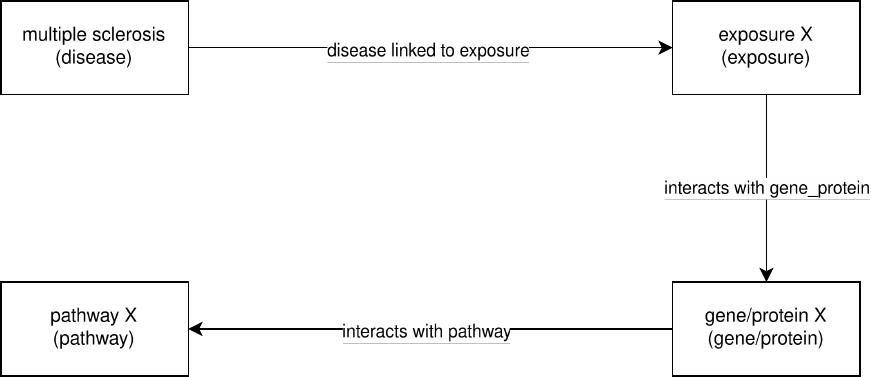}
  \caption{Example structure for 3-hop question (structure 4)}
  \label{fig:q3_s1}
\end{figure}
\item Structure 5: An entity has relationships interfacing with two other entities, one of which interfaces with a fourth entity, see \autoref{fig:q3_s2}. The question for this path was \textit{Which biological processes are affected by the gene APOE which are also affected by an exposure to something that is linked to multiple sclerosis?}
\begin{figure}
  \centering
  \includegraphics[scale=0.7]{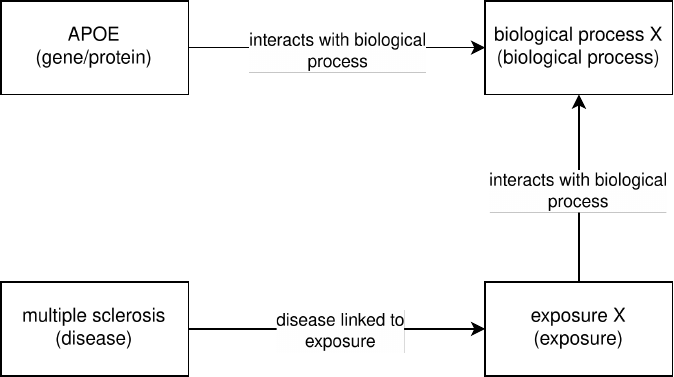}
  \caption{Example structure for 3-hop question (structure 5)}
  \label{fig:q3_s2}
\end{figure}
\end{itemize}

\section{Experimental Setup}
The LLMs used in our experiments are detailed in \autoref{llm_overview}. The core pipeline is built using the LangChain Expression Language (LCEL) \cite{lcel}. GPT models were called via their official APIs (prices at experiment time: GPT-4o \$2.50/1M tokens input and \$10/1M tokens output; GPT-4 Turbo \$10/1M tokens input and \$30/1M tokens output; GPT-5 \$1.25/1M tokens input and \$10/1M tokens output). Open-source models ran on our in-house cluster via Ollama and incurred no external billing. Wall-clock runtime was not benchmarked because latency depends on deployment choices (such as hardware allocation and batching), which are outside this study’s scope focused on correctness and the checker ablation.

\begin{table}[h!]
  \centering
  \caption{Overview of the used LLMs}
  \label{llm_overview}
  \begin{tabular}{lcll}
    \toprule
    \textbf{LLM} & \textbf{Number of Parameters} & \textbf{Open / Closed Source} & \textbf{Context length} \\
    \midrule
    bakllava:7b \cite{liu2023llava}            & 7b       & open & 32K  \\
    dolphin-llama3:8b \cite{dolphin_llama3}     & 8b       & open & 8K  \\
    dolphin-llama3:70b \cite{dolphin_llama3}    & 70b      & open & 8K  \\
    falcon2:11b \cite{malartic2024falcon2}           & 11b      & open  & 2K \\
    gemma2:9b \cite{gemma_2024}              & 9b       & open & 8K  \\
    goliath:120b \cite{goliath}                & 120b     & open & 4K  \\
    gpt-4 turbo  \cite{openai2023gpt4}          & unknown  & closed & 128K \\
    gpt-4o   \cite{openai2023gpt4turbo}              & unknown  & closed & 128K \\
    gpt-5 \cite{openai_introducing_gpt5_2025} & unknown & closed & 400K\\
    llama2:70b \cite{touvron2023llama}            & 70b      & open  & 4K \\
    llama3-chatqa:70b \cite{liu2024chatqa}     & 70b      & open & 8K  \\
    llama3:70b \cite{dubey2024llama}            & 70b      & open & 8K  \\
    llama3:8b  \cite{dubey2024llama}             & 8b       & open & 8K  \\
    llama3.3:70b \cite{meta_llama_3_3_70b_instruct_2024} & 70b & open & 128K \\
    medllama2:7b  \cite{medllama2}         & 7b       & open  & 4K \\
    mistral:7b \cite{jiang2023mistral}            & 7b       & open & 32K  \\
    mixtral:8x7b  \cite{jiang2024mixtral}         & 8x7b     & open & 32K  \\
    orca-mini:70b \cite{orca-mini}         & 70b      & open  & 4K \\
    qwen:32b  \cite{qwen}             & 32b      & open  & 32K \\
    qwen:110b   \cite{qwen}           & 110b     & open & 32K  \\
    starling-lm:7b    \cite{starling2023}     & 7b       & open & 8K  \\
    vicuna:33b      \cite{vicuna}       & 33b      & open & 2K  \\
    wizardlm2:7b   \cite{xu2023wizardlm}        & 7b       & open  & 32K \\
    \bottomrule
  \end{tabular}
\end{table}

\subsection{Evaluation Metrics}
We evaluated each LLM based on the number of correct answers across all 50 questions (\autoref{questions}) in our dataset. For each question we fixed a ground-truth result set during benchmark construction. Assuming the KG is correct, a model is correct if its generated Cypher, executed verbatim on the same database snapshot, returns exactly the ground-truth set of results (order ignored). 

\section{Cypher Generation with Zero-shot Prompting}
\label{llm_comparison}

In this experiment, we evaluated several LLMs' abilities to generate Cypher queries from natural language inputs in a zero-shot setting, where no examples or demonstrations are provided to guide the LLM. Each LLM was prompted to generate Cypher queries for the 50 questions in our benchmark set (see \autoref{questions}), with schema information provided as described in \autoref{cypher_query_gen}. The shortened prompt was adapted from the LangChain Cypher Graph QA chain \cite{cypher_langchain}. The full prompt can be found in \autoref{zero_shot}. The placeholders {schema} and {question} are replaced by the respective content before execution.

(Shortened) Zero-shot prompt:
\begin{verbatim}
Task: Generate Cypher statement to query a graph database.
Schema: {schema}
The question is: {question}
\end{verbatim}

To maintain consistency, each LLM had a temperature setting of 0 to reduce randomness, unless a custom temperature was not allowed. The generated Cypher queries were parsed to retain only relevant portions, removing non-Cypher content, and then processed by the query-checking algorithm (see \autoref{query_checker}) to correct common errors, including incorrect node types, relationship misdirections, and missing attributes.

\subsection{Results}
The proprietary GPT models, especially GPT-4 Turbo, often outperform the open-source models. \autoref{table:correct_table} reports, for each LLM, the number correct out of 50, accuracy, and 95\% Wilson CIs. \autoref{tab:result_mcnemar} shows Holm-adjusted McNemar pairwise tests, with significant results at \(\alpha=0.05\) bold and starred. According to these results, GPT-4 Turbo and GPT-5 outperform all open-source models except Llama 3.3 70B, where the difference is not significant. GPT-4o also does not differ significantly from Llama 3 70B. The GPT models do not differ significantly among themselves. Within the open-source group, Llama 3.3 70B leads, outperforming 13 models, followed by Llama 3 70B, which outperforms 9.

\begin{table}[htbp]
    \centering
    \caption{The number of correct answers (out of 50 total), accuracy and Wilson 95\% confidence intervals for all LLMs.}
    \label{table:correct_table}
    \begin{tabular}{l c c c}
        \toprule
        \textbf{LLM} & \textbf{\#correct} & \textbf{accuracy} & \textbf{Wilson 95\% CI} \\
        \midrule
        bakllava:7b & 0 & 0 & [0 , 0.071] \\ 
        medllama2:7b & 0 & 0 & [0 , 0.071] \\ 
        mistral:7b & 9 & 0.18 & [0.098 , 0.31] \\
        starling-lm:7b & 0 & 0 & [0 , 0.071] \\
        wizardlm2:7b & 5 & 0.1 &  [0.043 , 0.21] \\
        dolphin-llama3:8b& 8 & 0.16 & [0.083 , 0.29] \\
        llama3:8b& 11 & 0.22 & [0.13 , 0.35] \\
        gemma2:9b& 0 & 0 & [0 , 0.071] \\
        falcon2:11b& 0 & 0 & [0 , 0.071] \\
        qwen:32b& 20 & 0.4 & [0.28 , 0.54] \\
        vicuna:33b& 1 & 0.02 & [0.0035 , 0.1] \\
        mixtral:8x7b& 1 & 0.02 & [0.0035 , 0.1] \\
        dolphin-llama3:70b& 16 & 0.32 & [0.21 , 0.46] \\
        llama2:70b& 11 & 0.22 & [0.13 , 0.35] \\
        llama3:70b& 23 & 0.46 & [0.33 , 0.6] \\
        llama3.3:70b & 31 & 0.62 & [0.48, 0.74] \\
        llama3-chatqa:70b& 8 & 0.16 & [0.083 , 0.29] \\
        orca-mini:70b& 15 & 0.3 & [0.19 , 0.44] \\
        qwen:110b& 18 & 0.36 & [0.24 , 0.5] \\
        goliath:120b& 19 & 0.38 & [0.26 , 0.52] \\
        gpt-4 turbo& 45 & 0.9 &  [0.79, 0.96] \\
        gpt-4o& 40 & 0.8 & [0.67 , 0.89] \\
        gpt-5 & 42 & 0.84 & [0.74 , 0.94]\\
        \bottomrule
    \end{tabular}
\end{table}

\begin{table}[htbp]
\centering
\caption{Holm-adjusted McNemar pairwise comparisons of models (row-column); Green = Positive, Red = Negative. Bold and starred cells are significant (\(\alpha=0.05\)).}
\label{tab:result_mcnemar}
\scriptsize
\setlength{\tabcolsep}{2pt}
\renewcommand{\arraystretch}{1.05}
\newcommand{\pos}[1]{\cellcolor{green!12}{#1}}
\newcommand{\negc}[1]{\cellcolor{red!12}{#1}}
\resizebox{\textwidth}{!}{%
\begin{tabular}{lrrrrrrrrrrrrrrrrrrrrrrr}
\toprule
 & llama3:8b & llama3:70b & starling\_lm:7b & gpt-5 & gpt4-turbo & dolphin\_llama3:70b & llama3\_chatqa:70b & wizardlm2:7b & mistral:7b & medllama2:7b & qwen:110b & llama3.3:70b & gemma:9b & dolphin\_llama3:8b & bakllava:7b & falcon2:11b & llama2:70b & mixtral:8x7b & vicuna:33b & gpt-4o & orca\_mini:70b & goliath:120b & qwen:32b \\
\midrule
\textbf{llama3:8b} & - & \negc{$-0.24$} & \pos{+0.22} & \negc{\textbf{-0.62}$^{*}$} & \negc{\textbf{-0.68}$^{*}$} & \negc{$-0.10$} & \pos{+0.06} & \pos{+0.12} & \pos{+0.04} & \pos{+0.22} & \negc{$-0.14$} & \negc{\textbf{-0.40}$^{*}$} & \pos{+0.22} & \pos{+0.06} & \pos{+0.22} & \pos{+0.22} & 0.00 & \pos{+0.20} & \pos{+0.20} & \negc{\textbf{-0.58}$^{*}$} & \negc{$-0.08$} & \negc{$-0.16$} & \negc{$-0.18$} \\
\textbf{llama3:70b} & \pos{+0.24} & - & \pos{\textbf{+0.46}$^{*}$} & \negc{\textbf{-0.38}$^{*}$} & \negc{\textbf{-0.44}$^{*}$} & \pos{+0.14} & \pos{+0.30} & \pos{\textbf{+0.36}$^{*}$} & \pos{+0.28} & \pos{\textbf{+0.46}$^{*}$} & \pos{+0.10} & \negc{$-0.16$} & \pos{\textbf{+0.46}$^{*}$} & \pos{\textbf{+0.30}$^{*}$} & \pos{\textbf{+0.46}$^{*}$} & \pos{\textbf{+0.46}$^{*}$} & \pos{+0.24} & \pos{\textbf{+0.44}$^{*}$} & \pos{\textbf{+0.44}$^{*}$} & \negc{$-0.34$} & \pos{+0.16} & \pos{+0.08} & \pos{+0.06} \\
\textbf{starling\_lm:7b} & \negc{$-0.22$} & \negc{\textbf{-0.46}$^{*}$} & - & \negc{\textbf{-0.84}$^{*}$} & \negc{\textbf{-0.90}$^{*}$} & \negc{\textbf{-0.32}$^{*}$} & \negc{$-0.16$} & \negc{$-0.10$} & \negc{$-0.18$} & 0.00 & \negc{\textbf{-0.36}$^{*}$} & \negc{\textbf{-0.62}$^{*}$} & 0.00 & \negc{$-0.16$} & 0.00 & 0.00 & \negc{$-0.22$} & \negc{$-0.02$} & \negc{$-0.02$} & \negc{\textbf{-0.80}$^{*}$} & \negc{\textbf{-0.30}$^{*}$} & \negc{\textbf{-0.38}$^{*}$} & \negc{\textbf{-0.40}$^{*}$} \\
\textbf{gpt-5} & \pos{\textbf{+0.62}$^{*}$} & \pos{\textbf{+0.38}$^{*}$} & \pos{\textbf{+0.84}$^{*}$} & - & \negc{$-0.06$} & \pos{\textbf{+0.52}$^{*}$} & \pos{\textbf{+0.68}$^{*}$} & \pos{\textbf{+0.74}$^{*}$} & \pos{\textbf{+0.66}$^{*}$} & \pos{\textbf{+0.84}$^{*}$} & \pos{\textbf{+0.48}$^{*}$} & \pos{+0.22} & \pos{\textbf{+0.84}$^{*}$} & \pos{\textbf{+0.68}$^{*}$} & \pos{\textbf{+0.84}$^{*}$} & \pos{\textbf{+0.84}$^{*}$} & \pos{\textbf{+0.62}$^{*}$} & \pos{\textbf{+0.82}$^{*}$} & \pos{\textbf{+0.82}$^{*}$} & \pos{+0.04} & \pos{\textbf{+0.54}$^{*}$} & \pos{\textbf{+0.46}$^{*}$} & \pos{\textbf{+0.44}$^{*}$} \\
\textbf{gpt4-turbo} & \pos{\textbf{+0.68}$^{*}$} & \pos{\textbf{+0.44}$^{*}$} & \pos{\textbf{+0.90}$^{*}$} & \pos{+0.06} & - & \pos{\textbf{+0.58}$^{*}$} & \pos{\textbf{+0.74}$^{*}$} & \pos{\textbf{+0.80}$^{*}$} & \pos{\textbf{+0.72}$^{*}$} & \pos{\textbf{+0.90}$^{*}$} & \pos{\textbf{+0.54}$^{*}$} & \pos{+0.28} & \pos{\textbf{+0.90}$^{*}$} & \pos{\textbf{+0.74}$^{*}$} & \pos{\textbf{+0.90}$^{*}$} & \pos{\textbf{+0.90}$^{*}$} & \pos{\textbf{+0.68}$^{*}$} & \pos{\textbf{+0.88}$^{*}$} & \pos{\textbf{+0.88}$^{*}$} & \pos{+0.10} & \pos{\textbf{+0.60}$^{*}$} & \pos{\textbf{+0.52}$^{*}$} & \pos{\textbf{+0.50}$^{*}$} \\
\textbf{dolphin\_llama3:70b} & \pos{+0.10} & \negc{$-0.14$} & \pos{\textbf{+0.32}$^{*}$} & \negc{\textbf{-0.52}$^{*}$} & \negc{\textbf{-0.58}$^{*}$} & - & \pos{+0.16} & \pos{+0.22} & \pos{+0.14} & \pos{\textbf{+0.32}$^{*}$} & \negc{$-0.04$} & \negc{$-0.30$} & \pos{\textbf{+0.32}$^{*}$} & \pos{+0.16} & \pos{\textbf{+0.32}$^{*}$} & \pos{\textbf{+0.32}$^{*}$} & \pos{+0.10} & \pos{\textbf{+0.30}$^{*}$} & \pos{\textbf{+0.30}$^{*}$} & \negc{\textbf{-0.48}$^{*}$} & \pos{+0.02} & \negc{$-0.06$} & \negc{$-0.08$} \\
\textbf{llama3\_chatqa:70b} & \negc{$-0.06$} & \negc{$-0.30$} & \pos{+0.16} & \negc{\textbf{-0.68}$^{*}$} & \negc{\textbf{-0.74}$^{*}$} & \negc{$-0.16$} & - & \pos{+0.06} & \negc{$-0.02$} & \pos{+0.16} & \negc{$-0.20$} & \negc{\textbf{-0.46}$^{*}$} & \pos{+0.16} & 0.00 & \pos{+0.16} & \pos{+0.16} & \negc{$-0.06$} & \pos{+0.14} & \pos{+0.14} & \negc{\textbf{-0.64}$^{*}$} & \negc{$-0.14$} & \negc{$-0.22$} & \negc{$-0.24$} \\
\textbf{wizardlm2:7b} & \negc{$-0.12$} & \negc{\textbf{-0.36}$^{*}$} & \pos{+0.10} & \negc{\textbf{-0.74}$^{*}$} & \negc{\textbf{-0.80}$^{*}$} & \negc{$-0.22$} & \negc{$-0.06$} & - & \negc{$-0.08$} & \pos{+0.10} & \negc{\textbf{-0.26}$^{*}$} & \negc{\textbf{-0.52}$^{*}$} & \pos{+0.10} & \negc{$-0.06$} & \pos{+0.10} & \pos{+0.10} & \negc{$-0.12$} & \pos{+0.08} & \pos{+0.08} & \negc{\textbf{-0.70}$^{*}$} & \negc{$-0.20$} & \negc{$-0.28$} & \negc{$-0.30$} \\
\textbf{mistral:7b} & \negc{$-0.04$} & \negc{$-0.28$} & \pos{+0.18} & \negc{\textbf{-0.66}$^{*}$} & \negc{\textbf{-0.72}$^{*}$} & \negc{$-0.14$} & \pos{+0.02} & \pos{+0.08} & - & \pos{+0.18} & \negc{$-0.18$} & \negc{\textbf{-0.44}$^{*}$} & \pos{+0.18} & \pos{+0.02} & \pos{+0.18} & \pos{+0.18} & \negc{$-0.04$} & \pos{+0.16} & \pos{+0.16} & \negc{\textbf{-0.62}$^{*}$} & \negc{$-0.12$} & \negc{$-0.20$} & \negc{$-0.22$} \\
\textbf{medllama2:7b} & \negc{$-0.22$} & \negc{\textbf{-0.46}$^{*}$} & 0.00 & \negc{\textbf{-0.84}$^{*}$} & \negc{\textbf{-0.90}$^{*}$} & \negc{\textbf{-0.32}$^{*}$} & \negc{$-0.16$} & \negc{$-0.10$} & \negc{$-0.18$} & - & \negc{\textbf{-0.36}$^{*}$} & \negc{\textbf{-0.62}$^{*}$} & 0.00 & \negc{$-0.16$} & 0.00 & 0.00 & \negc{$-0.22$} & \negc{$-0.02$} & \negc{$-0.02$} & \negc{\textbf{-0.80}$^{*}$} & \negc{\textbf{-0.30}$^{*}$} & \negc{\textbf{-0.38}$^{*}$} & \negc{\textbf{-0.40}$^{*}$} \\
\textbf{qwen:110b} & \pos{+0.14} & \negc{$-0.10$} & \pos{\textbf{+0.36}$^{*}$} & \negc{\textbf{-0.48}$^{*}$} & \negc{\textbf{-0.54}$^{*}$} & \pos{+0.04} & \pos{+0.20} & \pos{\textbf{+0.26}$^{*}$} & \pos{+0.18} & \pos{\textbf{+0.36}$^{*}$} & - & \negc{$-0.26$} & \pos{\textbf{+0.36}$^{*}$} & \pos{+0.20} & \pos{\textbf{+0.36}$^{*}$} & \pos{\textbf{+0.36}$^{*}$} & \pos{+0.14} & \pos{\textbf{+0.34}$^{*}$} & \pos{\textbf{+0.34}$^{*}$} & \negc{\textbf{-0.44}$^{*}$} & \pos{+0.06} & \negc{$-0.02$} & \negc{$-0.04$} \\
\textbf{llama3.3:70b} & \pos{\textbf{+0.40}$^{*}$} & \pos{+0.16} & \pos{\textbf{+0.62}$^{*}$} & \negc{$-0.22$} & \negc{$-0.28$} & \pos{+0.30} & \pos{\textbf{+0.46}$^{*}$} & \pos{\textbf{+0.52}$^{*}$} & \pos{\textbf{+0.44}$^{*}$} & \pos{\textbf{+0.62}$^{*}$} & \pos{+0.26} & - & \pos{\textbf{+0.62}$^{*}$} & \pos{\textbf{+0.46}$^{*}$} & \pos{\textbf{+0.62}$^{*}$} & \pos{\textbf{+0.62}$^{*}$} & \pos{\textbf{+0.40}$^{*}$} & \pos{\textbf{+0.60}$^{*}$} & \pos{\textbf{+0.60}$^{*}$} & \negc{$-0.18$} & \pos{+0.32} & \pos{+0.24} & \pos{+0.22} \\
\textbf{gemma:9b} & \negc{$-0.22$} & \negc{\textbf{-0.46}$^{*}$} & 0.00 & \negc{\textbf{-0.84}$^{*}$} & \negc{\textbf{-0.90}$^{*}$} & \negc{\textbf{-0.32}$^{*}$} & \negc{$-0.16$} & \negc{$-0.10$} & \negc{$-0.18$} & 0.00 & \negc{\textbf{-0.36}$^{*}$} & \negc{\textbf{-0.62}$^{*}$} & - & \negc{$-0.16$} & 0.00 & 0.00 & \negc{$-0.22$} & \negc{$-0.02$} & \negc{$-0.02$} & \negc{\textbf{-0.80}$^{*}$} & \negc{\textbf{-0.30}$^{*}$} & \negc{\textbf{-0.38}$^{*}$} & \negc{\textbf{-0.40}$^{*}$} \\
\textbf{dolphin\_llama3:8b} & \negc{$-0.06$} & \negc{\textbf{-0.30}$^{*}$} & \pos{+0.16} & \negc{\textbf{-0.68}$^{*}$} & \negc{\textbf{-0.74}$^{*}$} & \negc{$-0.16$} & 0.00 & \pos{+0.06} & \negc{$-0.02$} & \pos{+0.16} & \negc{$-0.20$} & \negc{\textbf{-0.46}$^{*}$} & \pos{+0.16} & - & \pos{+0.16} & \pos{+0.16} & \negc{$-0.06$} & \pos{+0.14} & \pos{+0.14} & \negc{\textbf{-0.64}$^{*}$} & \negc{$-0.14$} & \negc{$-0.22$} & \negc{$-0.24$} \\
\textbf{bakllava:7b} & \negc{$-0.22$} & \negc{\textbf{-0.46}$^{*}$} & 0.00 & \negc{\textbf{-0.84}$^{*}$} & \negc{\textbf{-0.90}$^{*}$} & \negc{\textbf{-0.32}$^{*}$} & \negc{$-0.16$} & \negc{$-0.10$} & \negc{$-0.18$} & 0.00 & \negc{\textbf{-0.36}$^{*}$} & \negc{\textbf{-0.62}$^{*}$} & 0.00 & \negc{$-0.16$} & - & 0.00 & \negc{$-0.22$} & \negc{$-0.02$} & \negc{$-0.02$} & \negc{\textbf{-0.80}$^{*}$} & \negc{\textbf{-0.30}$^{*}$} & \negc{\textbf{-0.38}$^{*}$} & \negc{\textbf{-0.40}$^{*}$} \\
\textbf{falcon2:11b} & \negc{$-0.22$} & \negc{\textbf{-0.46}$^{*}$} & 0.00 & \negc{\textbf{-0.84}$^{*}$} & \negc{\textbf{-0.90}$^{*}$} & \negc{\textbf{-0.32}$^{*}$} & \negc{$-0.16$} & \negc{$-0.10$} & \negc{$-0.18$} & 0.00 & \negc{\textbf{-0.36}$^{*}$} & \negc{\textbf{-0.62}$^{*}$} & 0.00 & \negc{$-0.16$} & 0.00 & - & \negc{$-0.22$} & \negc{$-0.02$} & \negc{$-0.02$} & \negc{\textbf{-0.80}$^{*}$} & \negc{\textbf{-0.30}$^{*}$} & \negc{\textbf{-0.38}$^{*}$} & \negc{\textbf{-0.40}$^{*}$} \\
\textbf{llama2:70b} & 0.00 & \negc{$-0.24$} & \pos{+0.22} & \negc{\textbf{-0.62}$^{*}$} & \negc{\textbf{-0.68}$^{*}$} & \negc{$-0.10$} & \pos{+0.06} & \pos{+0.12} & \pos{+0.04} & \pos{+0.22} & \negc{$-0.14$} & \negc{\textbf{-0.40}$^{*}$} & \pos{+0.22} & \pos{+0.06} & \pos{+0.22} & \pos{+0.22} & - & \pos{+0.20} & \pos{+0.20} & \negc{\textbf{-0.58}$^{*}$} & \negc{$-0.08$} & \negc{$-0.16$} & \negc{$-0.18$} \\
\textbf{mixtral:8x7b} & \negc{$-0.20$} & \negc{\textbf{-0.44}$^{*}$} & \pos{+0.02} & \negc{\textbf{-0.82}$^{*}$} & \negc{\textbf{-0.88}$^{*}$} & \negc{\textbf{-0.30}$^{*}$} & \negc{$-0.14$} & \negc{$-0.08$} & \negc{$-0.16$} & \pos{+0.02} & \negc{\textbf{-0.34}$^{*}$} & \negc{\textbf{-0.60}$^{*}$} & \pos{+0.02} & \negc{$-0.14$} & \pos{+0.02} & \pos{+0.02} & \negc{$-0.20$} & - & 0.00 & \negc{\textbf{-0.78}$^{*}$} & \negc{\textbf{-0.28}$^{*}$} & \negc{\textbf{-0.36}$^{*}$} & \negc{\textbf{-0.38}$^{*}$} \\
\textbf{vicuna:33b} & \negc{$-0.20$} & \negc{\textbf{-0.44}$^{*}$} & \pos{+0.02} & \negc{\textbf{-0.82}$^{*}$} & \negc{\textbf{-0.88}$^{*}$} & \negc{\textbf{-0.30}$^{*}$} & \negc{$-0.14$} & \negc{$-0.08$} & \negc{$-0.16$} & \pos{+0.02} & \negc{\textbf{-0.34}$^{*}$} & \negc{\textbf{-0.60}$^{*}$} & \pos{+0.02} & \negc{$-0.14$} & \pos{+0.02} & \pos{+0.02} & \negc{$-0.20$} & 0.00 & - & \negc{\textbf{-0.78}$^{*}$} & \negc{$-0.28$} & \negc{\textbf{-0.36}$^{*}$} & \negc{\textbf{-0.38}$^{*}$} \\
\textbf{gpt-4o} & \pos{\textbf{+0.58}$^{*}$} & \pos{+0.34} & \pos{\textbf{+0.80}$^{*}$} & \negc{$-0.04$} & \negc{$-0.10$} & \pos{\textbf{+0.48}$^{*}$} & \pos{\textbf{+0.64}$^{*}$} & \pos{\textbf{+0.70}$^{*}$} & \pos{\textbf{+0.62}$^{*}$} & \pos{\textbf{+0.80}$^{*}$} & \pos{\textbf{+0.44}$^{*}$} & \pos{+0.18} & \pos{\textbf{+0.80}$^{*}$} & \pos{\textbf{+0.64}$^{*}$} & \pos{\textbf{+0.80}$^{*}$} & \pos{\textbf{+0.80}$^{*}$} & \pos{\textbf{+0.58}$^{*}$} & \pos{\textbf{+0.78}$^{*}$} & \pos{\textbf{+0.78}$^{*}$} & - & \pos{\textbf{+0.50}$^{*}$} & \pos{\textbf{+0.42}$^{*}$} & \pos{\textbf{+0.40}$^{*}$} \\
\textbf{orca\_mini:70b} & \pos{+0.08} & \negc{$-0.16$} & \pos{\textbf{+0.30}$^{*}$} & \negc{\textbf{-0.54}$^{*}$} & \negc{\textbf{-0.60}$^{*}$} & \negc{$-0.02$} & \pos{+0.14} & \pos{+0.20} & \pos{+0.12} & \pos{\textbf{+0.30}$^{*}$} & \negc{$-0.06$} & \negc{$-0.32$} & \pos{\textbf{+0.30}$^{*}$} & \pos{+0.14} & \pos{\textbf{+0.30}$^{*}$} & \pos{\textbf{+0.30}$^{*}$} & \pos{+0.08} & \pos{\textbf{+0.28}$^{*}$} & \pos{+0.28} & \negc{\textbf{-0.50}$^{*}$} & - & \negc{$-0.08$} & \negc{$-0.10$} \\
\textbf{goliath:120b} & \pos{+0.16} & \negc{$-0.08$} & \pos{\textbf{+0.38}$^{*}$} & \negc{\textbf{-0.46}$^{*}$} & \negc{\textbf{-0.52}$^{*}$} & \pos{+0.06} & \pos{+0.22} & \pos{+0.28} & \pos{+0.20} & \pos{\textbf{+0.38}$^{*}$} & \pos{+0.02} & \negc{$-0.24$} & \pos{\textbf{+0.38}$^{*}$} & \pos{+0.22} & \pos{\textbf{+0.38}$^{*}$} & \pos{\textbf{+0.38}$^{*}$} & \pos{+0.16} & \pos{\textbf{+0.36}$^{*}$} & \pos{\textbf{+0.36}$^{*}$} & \negc{\textbf{-0.42}$^{*}$} & \pos{+0.08} & - & \negc{$-0.02$} \\
\textbf{qwen:32b} & \pos{+0.18} & \negc{$-0.06$} & \pos{\textbf{+0.40}$^{*}$} & \negc{\textbf{-0.44}$^{*}$} & \negc{\textbf{-0.50}$^{*}$} & \pos{+0.08} & \pos{+0.24} & \pos{\textbf{+0.30}$^{*}$} & \pos{+0.22} & \pos{\textbf{+0.40}$^{*}$} & \pos{+0.04} & \negc{$-0.22$} & \pos{\textbf{+0.40}$^{*}$} & \pos{+0.24} & \pos{\textbf{+0.40}$^{*}$} & \pos{\textbf{+0.40}$^{*}$} & \pos{+0.18} & \pos{\textbf{+0.38}$^{*}$} & \pos{\textbf{+0.38}$^{*}$} & \negc{\textbf{-0.40}$^{*}$} & \pos{+0.10} & \pos{+0.02} & - \\
\bottomrule
\end{tabular}%
}
\end{table}

\begin{figure}
  \centering
  \includegraphics[scale=0.4]{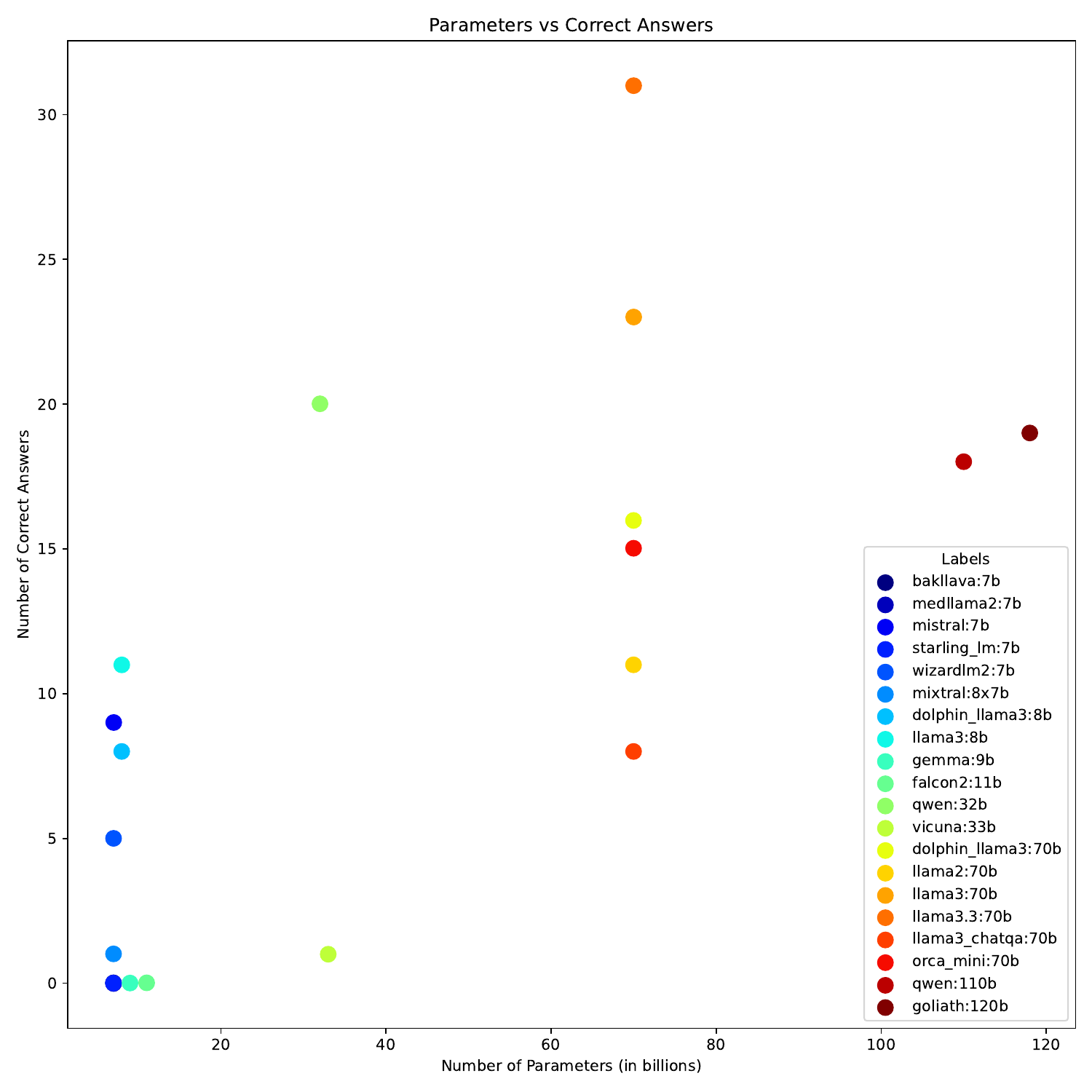}
  \caption{Scatterplot of number of parameters by correct answers; excluding GPT because the number of parameters is likely too high}
  \label{fig:scatterplot}
\end{figure}

\autoref{fig:scatterplot} depicts the relationship between the number of parameters (in billions) and the number of correct answers. Although proprietary GPT models were excluded due to undisclosed parameter counts, a general trend indicates that models with more parameters tend to yield better results.\

\begin{figure}
  \centering
  \includegraphics[width=0.9\textwidth]{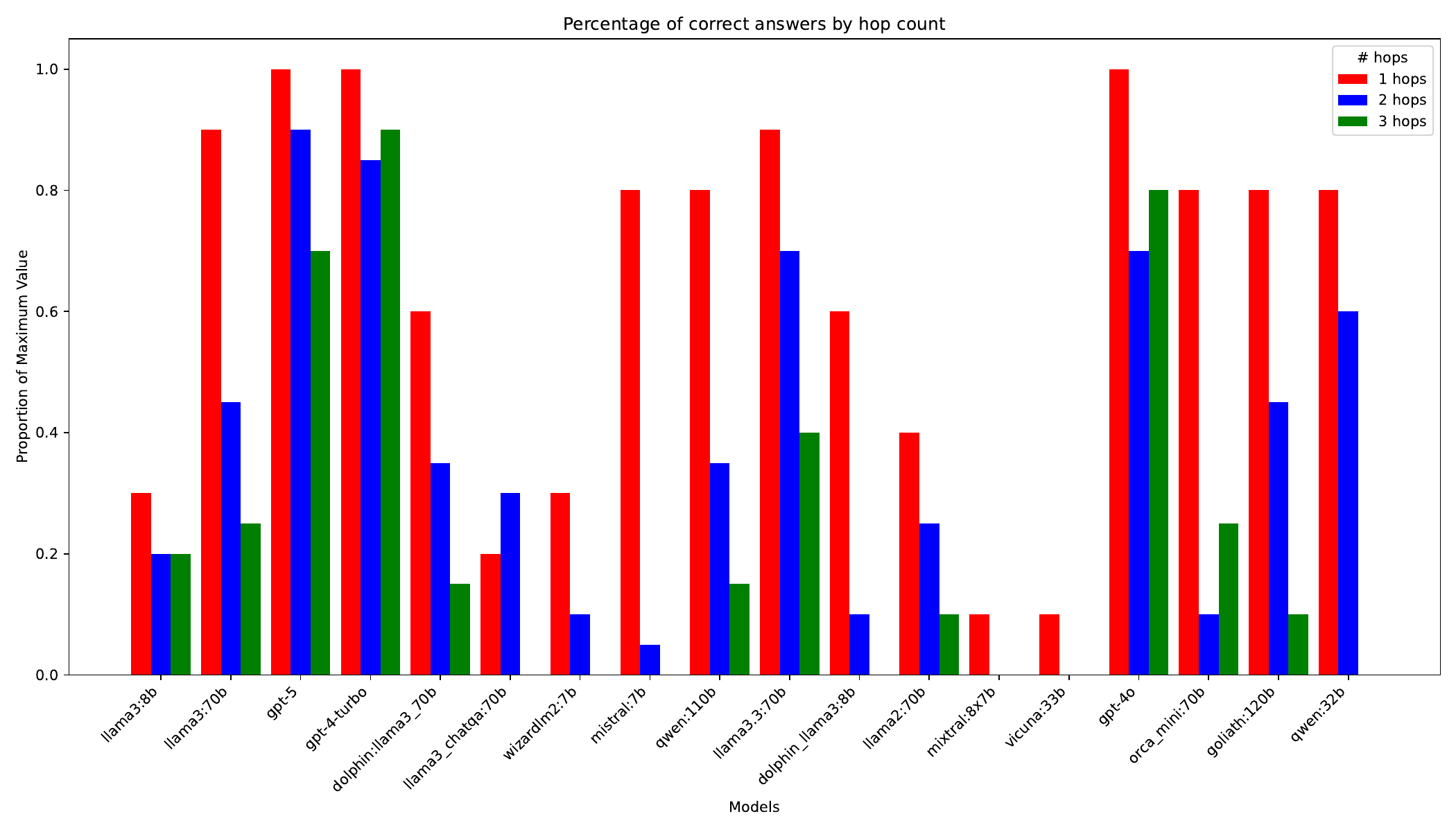}
  \caption{Percentage of correct answers by hop count}
  \label{fig:hop_plot}
\end{figure}

\autoref{fig:hop_plot} explores model performance based on query complexity, measured by the "hop" count between entities. Most models performed well on simpler, one-hop queries, but only the GPT models maintained high accuracy on both of the more complex two- and three-hop queries.

Overall, our findings indicate that for generating reliable Cypher queries from biomedical questions with a zero-shot prompt, the GPT variants are currently the most effective models within the proposed framework. Among them, GPT-4 Turbo and GPT-5 outperformed more open source variants than the GPT-4o model.

\subsection{Influence of the Query Checker}

We evaluate the checker by comparing, per model and item, correctness before vs after applying it. The checker is monotone by design: as a rule-based algorithm, it never alters a correct query, and either fixes an error or leaves it unchanged. Across all 23×50 evaluations it corrected 172 of 999 items that were wrong pre-checker, for an overall help rate 0.172 with a Wilson 95\% CI of [0.15, 0.2] (Wilson 95\% CI reported per model in \autoref{tab:checker-ablation}). This yields an average accuracy uplift of +14.95 percentage points.

The effect is model-dependent. The checker amplifies strong models: GPT-5 shows a help rate of 0.84 [0.71, 0.91], GPT-4o 0.80 [0.66, 0.89], and GPT-4 Turbo 0.69 [0.44, 0.86], with sizable uplift and low residual error. Notably, without the checker the first two models have few correct queries, yet they rank among the highest once checking is applied. In contrast, Llama 3.3:70b starts with many correct queries but achieves only a 0.14 help rate, suggesting its remaining errors are harder for the checker to repair. Mid-tier models benefit but retain significant residual error, for example Llama 3:70b at 0.45 help and 0.54 residual, and Goliath:120b at 0.31 help and 0.62 residual. Low-baseline models rarely improve, often near 0 with Wilson upper bounds around 0.07. In conclusion, the checker helps most when the base model is already strong, gives moderate gains in the middle, and cannot rescue low-end failures.

The checker addresses schema/syntax issues (e.g., node names or relationship directions). It cannot repair conceptual errors or missing reasoning steps. Therefore it helps most when the model’s queries are “almost right” and least when queries are structurally off or syntactically incorrect.
A pooled one-sided binomial test on help rate confirms the checker does not fix a majority of errors overall (baseline p0=0.5), but it yields meaningful accuracy gains in aggregate due to the volume of eligible near-misses.

\begin{table}[t]
\centering
\small
\setlength{\tabcolsep}{4pt}
\renewcommand{\arraystretch}{1.1}
\resizebox{\linewidth}{!}{
\begin{tabular}{lrrrrrrrrlr}
\toprule
Model & \makecell{Correct w/o\\Checker} & \makecell{Corrected by\\Checker} & \makecell{Wrong, no \\checking opportunities} & \makecell{Wrong despite \\checking} & Total correct & Total false & \makecell{Accuracy\\uplift} & Help rate & \makecell{Help rate \\ 95\% Wilson CI} & \makecell{Residual\\error rate}\\
\midrule
llama3:8b & 6 & 5 & 14 & 25 & 11 & 39 & 0.10 & 0.11 &  [0.05, 0.24] & 0.78\\
llama3:70b & 1 & 22 & 8 & 19 & 23 & 27 & 0.44 & 0.45 & [0.32, 0.59] & 0.54 \\
starling\_lm:7b & 0 & 0 & 0 & 50 & 0 & 50 & 0.00 & 0.00 & [0, 0.071] & 1.00\\
gpt-4 turbo & 34 & 11 & 4 & 1 & 45 & 5 & 0.22 & 0.69 & [0.44, 0.86] & 0.10 \\
gpt-5 & 1 & 41 & 0 & 8 & 42 & 8 & 0.82 & 0.84 & [0.71, 0.91] & 0.16\\
dolphin\_llama3:70b & 10 & 6 & 8 & 26 & 16 & 34 & 0.12 & 0.15 & [0.071, 0.29] & 0.68\\
llama3\_chatqa:70b & 7 & 1 & 10 & 32 & 8 & 42 & 0.02 & 0.02 & [0.0041, 0.12] & 0.84 \\
wizardlm2:7b & 3 & 2 & 16 & 29 & 5 & 45 & 0.04 & 0.04 & [0.012, 0.14] & 0.90\\
mistral:7b & 7 & 2 & 16 & 25 & 9 & 41 & 0.04 & 0.05 & [0.013, 0.15] & 0.82\\
medllama2:7b & 0 & 0 & 19 & 31 & 0 & 50 & 0.00 & 0.00 & [0, 0.071] & 1.00\\
qwen:110b & 13 & 5 & 10 & 22 & 18 & 32 & 0.10 & 0.14 & [0.059, 0.28] & 0.64\\
gemma2:9b & 0 & 0 & 0 & 50 & 0 & 50 & 0.00 & 0.00 & [0, 0.071] & 1.00\\
dolphin\_llama3:8b & 7 & 1 & 29 & 13 & 8 & 42 & 0.02 & 0.02 & [0.0041, 0.12] & 0.84\\
bakllava:7b & 0 & 0 & 0 & 50 & 0 & 50 & 0.00 & 0.00 & [0, 0.071] & 1.00\\
falcon2:11b & 0 & 0 & 0 & 50 & 0 & 50 & 0.00 & 0.00 & [0, 0.071] & 1.00\\
llama2:70b & 10 & 1 & 14 & 25 & 11 & 39 & 0.02 & 0.02 & [0.0044, 0.13] & 0.78\\
mixtral:8x7b & 1 & 0 & 1 & 48 & 1 & 49 & 0.00 & 0.00 & [0, 0.073] & 0.98 \\
vicuna:33b & 0 & 1 & 23 & 26 & 1 & 49 & 0.02 & 0.02 & [0.0035, 0.1] & 0.98\\
gpt-4o & 1 & 39 & 0 & 10 & 40 & 10 & 0.78 & 0.80 & [0.66, 0.89] & 0.20\\
orca\_mini:70b & 3 & 12 & 6 & 29 & 15 & 35 & 0.24 & 0.26 & [0.15, 0.4] & 0.70\\
goliath:120b & 5 & 14 & 13 & 18 & 19 & 31 & 0.28 & 0.31 & [0.2, 0.46] & 0.62\\
qwen:32b & 14 & 6 & 13 & 17 & 20 & 30 & 0.12 & 0.17 & [0.079, 0.32] & 0.60\\
llama3.3:70b & 28 & 3 & 8 & 11 & 31 & 19 & 0.06 & 0.14 & [0.047, 0.33] & 0.38\\
\bottomrule
\end{tabular}}
\caption{Per-model pre/post query checker results on 50 items per model. Counts show items correct without the checker, corrected by the checker, still wrong (no checking opportunities or despite checking), and totals. Accuracy uplift is defined as the increase in accuracy when applying the checker. Help rate is the effectiveness on eligible errors with 95\% Wilson CIs. Residual error rate is the remaining error rate after the checker.}
\label{tab:checker-ablation}
\end{table}

\subsection{Influence of Paraphrased Questions}
We tested whether paraphrasing affects pipeline performance by rewriting all 50 questions from \autoref{questions} (\autoref{sec:paraphrased}) and re-evaluating the two top models: GPT-4 Turbo (closed-source) and Llama 3.3 70B (open-source). \autoref{table:paraphrased_accuracy} reports the number of correct answers, accuracy, and 95\% Wilson CIs. Using McNemar tests with Holm correction, \autoref{tab:paraphrased_result_mcnemar} compares each model to itself across normal vs paraphrased questions and compares the models to each other on each set. We find no significant within-model difference between normal and paraphrased questions. On the normal set, GPT-4 Turbo significantly outperforms Llama 3.3 70B; this contrast was not flagged in \autoref{tab:result_mcnemar} because that table applies a more stringent correction across a larger family of comparisons. Finally, \autoref{tab:paraphrased_total_result_mcnemar} aggregates normal and paraphrased questions and shows that GPT-4 Turbo remains significantly better than Llama 3.3 70B; total answer counts per model in this combined analysis are also reported in \autoref{table:paraphrased_accuracy}.

\begin{table}[htbp]
    \centering
    \caption{The number of correct answers, accuracy and Wilson 95\% confidence intervals for all LLMs.}
    \label{table:paraphrased_accuracy}
    \begin{tabular}{l c c c c c}
        \toprule
        \textbf{LLM} & \textbf{Question Type} & \textbf{\#Correct} & \textbf{\#Total} & \textbf{Accuracy} & \textbf{Wilson 95\% CI} \\
        \midrule
        gpt-4 turbo & normal & 45 & 50 & 0.9 & [0.79, 0.96]\\ 
        llama3.3:70b & normal & 31 & 50 &0.62  &[0.48, 0.74] \\ 
        gpt-4 turbo & paraphrased & 41 & 50 & 0.82& [0.69, 0.9]\\ 
        llama3.3:70b & paraphrased & 32 & 50 & 0.64& [0.5, 0.76]\\ 
        gpt-4 turbo & both & 86 & 100 &0.86 & [0.78, 0.91] \\ 
        llama3.3:70b & both & 63& 100 & 0.63&[0.54 , 0.72] \\ 
        \bottomrule
    \end{tabular}
\end{table}

\begin{table}[htbp]
\centering
\caption{Holm-adjusted McNemar pairwise comparisons of GPT-4 Turbo and Llama 3.3:70b on both normal and paraphrased questions (row-column); Green = Positive, Red = Negative. Bold and starred cells are significant (\(\alpha=0.05\)).}
\label{tab:paraphrased_result_mcnemar}
\scriptsize
\setlength{\tabcolsep}{2pt}
\renewcommand{\arraystretch}{1.05}
\newcommand{\pos}[1]{\cellcolor{green!12}{#1}}
\newcommand{\negc}[1]{\cellcolor{red!12}{#1}}
\resizebox{\textwidth}{!}{%
\begin{tabular}{lrrrr}
\toprule
 & gpt-4 turbo normal & llama3.3:70b normal & gpt-4 turbo paraphrased & llama3.3:70b paraphrased\\
\midrule
\textbf{gpt-4 turbo normal} & - & \pos{\textbf{+0.28}$^{*}$} & \pos{+0.08} & \pos{\textbf{+0.26}$^{*}$} \\
\textbf{llama3.3:70b normal} & \negc{\textbf{-0.28}$^{*}$} & - & \negc{$-0.20$} & \negc{$-0.02$} \\
\textbf{gpt-4 turbo paraphrased} & \negc{$-0.08$} & \pos{+0.20} & - & \pos{+0.18} \\
\textbf{llama3.3:70b paraphrased} & \negc{\textbf{-0.26}$^{*}$} & \pos{+0.02} & \negc{$-0.18$} & - \\
\bottomrule
\end{tabular}%
}
\end{table}

\begin{table}[htbp]
\centering
\caption{Holm-adjusted McNemar pairwise comparisons of GPT-4 Turbo and Llama 3.3:70b on normal and paraphrased questions together (row-column); Green = Positive, Red = Negative. Bold and starred cells are significant (\(\alpha=0.05\)).}
\label{tab:paraphrased_total_result_mcnemar}
\scriptsize
\setlength{\tabcolsep}{2pt}
\renewcommand{\arraystretch}{1.05}
\newcommand{\pos}[1]{\cellcolor{green!12}{#1}}
\newcommand{\negc}[1]{\cellcolor{red!12}{#1}}
\begin{tabular}{lrr}
\toprule
 & gpt-4 turbo& llama3.3:70b\\
\midrule
\textbf{gpt-4 turbo} & - & \pos{\textbf{+0.23}$^{}$}  \\
\textbf{llama3.3:70b} & \negc{\textbf{-0.23}$^{}$} & -  \\
\bottomrule
\end{tabular}%

\end{table}

\section{Cypher Generation with Optimized Prompting} 
\label{optimization}

In this experiment, we investigated how prompt optimization could improve the quality of Cypher queries generated by LLMs. Using the 50 questions from our benchmark dataset, we tested optimized prompts, including specific examples and tailored instructions, to help the LLMs produce more accurate Cypher queries.

This section presents results from experiments focused on optimizing prompts for the two models GPT-4 Turbo and Llama 3:70b. We tested three prompting strategies (zero-shot, one-shot, and few-shot prompting) and explored various prompt crafting techniques to enhance initial Cypher query generation and improve overall pipeline performance.

\subsection{Multi-shot Prompts} 
\label{shot_prompting}

Providing different numbers of examples in prompts can significantly affect model performance. To evaluate this effect, we compared zero-shot (no examples), one-shot (one example), and few-shot (multiple examples) prompting to assess their impact on Cypher query generation. Below are examples of these prompts; full versions with more hints for the LLM are available in Appendix \autoref{one-shot} and Appendix \autoref{few_shot}.

(Shortened) One-Shot Prompt:
\begin{verbatim}
Task:Generate Cypher statement to query a graph database.

Schema: {schema}

Follow these Cypher example when Generating Cypher statements:
# How many actors played in Top Gun?
MATCH (m:movie {{name:"Top Gun"}})<-[:acted_in]-(a:actor)
RETURN a.name

The question is:
{question}
\end{verbatim}

(Shortened) Few-Shot Prompt:
\begin{verbatim}
Task:Generate Cypher statement to query a graph database.

Schema: {schema}

Follow these Cypher example when Generating Cypher statements:
# Which actors played in Top Gun?
MATCH (m:movie {{name:"Top Gun"}})<-[:acted_in]-(a:actor)
RETURN a.name
# What town were the actors that played in Top Gun born in?
MATCH (m:movie {{name:"Top Gun"}})<-[:acted_in]-(a:actor)-[:born_in]->(t:town)
RETURN t.name

The question is:
{question}
\end{verbatim}

\autoref{table:shot_table} shows the total number of correct answers out of a maximum of 50, the accuracy and the Wilson 95\% confidence intervals for GPT-4 Turbo and Llama 3:70b. When doing a Holm-adjusted McNemar pairwise comparison of the prompt types and models, as shown in \autoref{tab:mcnemar_nshot}, we see that for GPT-4 Turbo, adding examples does not significantly change performance. However, for Llama 3:70b, few-shot prompting showed a significant improvement over zero-shot. One-shot does not lead to significant improvements. When comparing Llama 3:70b and GPT-4 Turbo, combining the GPT model with zero-shot prompting significantly outperforms all Llama 3 variants except LLama 3 Few-Shot. Moreover, every shot-combination coupled with a GPT model outperforms Llama 3 zero-shot prompting.  This underscores the advantage of using few-shot prompting with a model like Llama 3:70b.

\begin{table}[htbp]
\setlength{\tabcolsep}{4pt}
\small
\centering
\caption{n-shot comparison for GPT-4 Turbo and Llama 3:70b: the number of correct answers (out of 50 total), accuracy and Wilson 95\% confidence intervals.}
\label{table:shot_table}
\begin{tabular}{l ccc ccc ccc}
\toprule
\textbf{LLM} &
\multicolumn{3}{c}{\textbf{Zero-shot}} &
\multicolumn{3}{c}{\textbf{One-shot}} &
\multicolumn{3}{c}{\textbf{Few-shot}} \\
\cmidrule(lr){2-4}\cmidrule(lr){5-7}\cmidrule(lr){8-10}
& \textbf{k} & $\hat{p}$ & \textbf{CI}
& \textbf{k} & $\hat{p}$ & \textbf{CI}
& \textbf{k} & $\hat{p}$ & \textbf{CI} \\
\midrule
gpt-4 turbo      & 45 & 0.9 & [0.79, 0.96] & 42 & 0.84 & [0.71 , 0.92] & 42 & 0.84 & [0.71 , 0.92] \\
llama3:70b  & 23 & 0.46 & [0.33 , 0.6] & 32 & 0.64 & [0.5 , 0.76] & 35 & 0.7 & [0.56 , 0.81] \\
\bottomrule
\end{tabular}
\end{table}

\begin{table}[htbp]
\centering
\caption{Holm-adjusted McNemar pairwise comparisons of Zero-Shot, One-Shot and Few-Shot prompts for GPT-4 Turbo and Llama 3:70b; Green = Positive, Red = Negative. Bold and starred cells are significant (\(\alpha=0.05\)).}
\label{tab:mcnemar_nshot}
\scriptsize
\setlength{\tabcolsep}{2pt}
\renewcommand{\arraystretch}{1.05}
\newcommand{\pos}[1]{\cellcolor{green!12}{#1}}
\newcommand{\negc}[1]{\cellcolor{red!12}{#1}}
\resizebox{\textwidth}{!}{%
\begin{tabular}{lrrrrrr}
\toprule
 & llama3:70b Zero-Shot & llama3:70b One-Shot & llama3:70b Few-Shot & gpt-4 turbo Zero-Shot & gpt-4 turbo One-Shot & gpt-4 turbo Few-Shot\\
\midrule
\textbf{llama3:70b Zero-Shot} & \textemdash
& \negc{$-0.18$} & \negc{$\boldsymbol{-0.24^{*}}$} & \negc{$\boldsymbol{-0.44^{*}}$} & \negc{$\boldsymbol{-0.38^{*}}$} & \negc{$\boldsymbol{-0.38^{*}}$} \\
\textbf{llama3:70b One-Shot} & \pos{$+0.18$}
& \textemdash & \negc{$-0.06$} & \negc{$\boldsymbol{-0.26^{*}}$} & \negc{$-0.20$} & \negc{$-0.20$} \\
\textbf{llama3:70b Few-Shot} & \pos{$\boldsymbol{+0.24^{*}}$}
& \pos{$+0.06$} & \textemdash & \negc{$-0.20$} & \negc{$-0.14$} & \negc{$-0.14$} \\
\textbf{gpt-4 turbo Zero-Shot} & \pos{$\boldsymbol{+0.44^{*}}$}
& \pos{$\boldsymbol{+0.26^{*}}$} & \pos{$+0.20$} & \textemdash & \pos{$+0.06$} & \pos{$+0.06$} \\
\textbf{gpt-4 turbo One-Shot} & \pos{$\boldsymbol{+0.38^{*}}$}
& \pos{$+0.20$} & \pos{$+0.14$} & \negc{$-0.06$} & \textemdash & 0.00 \\
\textbf{gpt-4 turbo Few-Shot} & \pos{$\boldsymbol{+0.38^{*}}$}
& \pos{$+0.20$} & \pos{$+0.14$} & \negc{$-0.06$} & 0.00 & \textemdash \\
\bottomrule
\end{tabular}%
}
\end{table}

\subsection{Promptcrafting} 
\label{promptcrafting}
Prompt design can substantially influence model performance. To explore this, we tested different prompt formulations to measure their impact on Cypher query generation. Building on our initial prompts, we experimented with several variations:
\begin{itemize}
\item Simplified Prompt: Removed detailed instructions to evaluate the effect of minimal guidance (Appendix \autoref{simple_prompt})
\item Syntax Emphasis Prompt: Added a sentence instructing the model to focus on correct syntax usage (Appendix \autoref{syntax_prompt})\ 
\item Social Engineering Prompt: Introduced a scenario where the prompter could face consequences (e.g., being fired) if the LLM made mistakes (Appendix \autoref{social_prompt})
\item Expert Role Prompt: Positioned the LLM as a Cypher query expert to encourage more accurate query generation (Appendix \autoref{role_prompt})
\end{itemize}

These variations were applied to both GPT-4 Turbo and Llama 3:70b. \autoref{table:shot_table} shows the total number of correct answers out of a maximum of 50, the accuracy and the Wilson 95\% confidence intervals for GPT-4 Turbo and Llama 3:70b. 
When doing a Holm-adjusted McNemar pairwise comparison of the prompt types and models, as shown in \autoref{tab:mcnemar_promptcrafting}, we see that the simplified prompt was the worst choice for GPT, with the Social and Expert prompts being significantly better. For Llama 3:70b, the promptcrafting did not have a statistically significant impact on performance. When comparing GPT-4 Turbo and Llama 3:70b, it is notable that the GPT model performs significantly better, except for the Simple prompt, which only performs significantly better than Llama 3 combined with the same prompt.

\begin{table}[htbp]
\setlength{\tabcolsep}{4pt}
\small
\centering
\caption{Promptcrafting comparison for GPT-4 Turbo and Llama 3:70b: the number of correct answers (out of 50 total), accuracy and Wilson 95\% confidence intervals. The compared prompts are the standard prompt used in the experiment above, a simplified prompt with minimal guidance, a prompt with special emphasis on correct syntax, a social engineering prompt and a prompt positioning the LLM as an expert.}
\label{table:prompt_comparison_table}
\begin{tabular}{l ccc ccc}
\toprule
\textbf{Prompt} &
\multicolumn{3}{c}{\textbf{gpt-4 turbo}} &
\multicolumn{3}{c}{\textbf{llama3:70b}} \\
\cmidrule(lr){2-4}\cmidrule(lr){5-7}
& \textbf{k} & $\hat{p}$ & \textbf{CI} & \textbf{k} & $\hat{p}$ & \textbf{CI} \\
\midrule
Standard           & 45 & 0.90 & [0.79, 0.96] & 23 & 0.46 &  [0.33 , 0.6] \\
Simplified         & 34 & 0.68 & [0.54 , 0.79] & 19 & 0.38 & [0.26 , 0.52] \\
Syntax Emphasis    & 45 & 0.9 &  [0.79 , 0.96] & 24 & 0.48 & [0.35 , 0.61] \\
Social Engineering & 46 & 0.92 & [0.81 , 0.97] & 29 & 0.58 & [0.44 , 0.71] \\
Expert Role        & 46 & 0.92 & [0.81 , 0.97] & 26 & 0.52 &  [0.39 , 0.65]\\
\bottomrule
\end{tabular}
\end{table}

\begin{table}[htbp]
\centering
\caption{Holm-adjusted McNemar pairwise comparisons of different prompt types for the best GPT-model (GPT-4 Turbo) and the best open source model (Llama 3:70b); Green = Positive, Red = Negative. Bold and starred cells are significant (\(\alpha=0.05\)).}
\label{tab:mcnemar_promptcrafting}
\scriptsize
\setlength{\tabcolsep}{2pt}
\renewcommand{\arraystretch}{1.05}
\newcommand{\pos}[1]{\cellcolor{green!12}{#1}}
\newcommand{\negc}[1]{\cellcolor{red!12}{#1}}
\resizebox{\textwidth}{!}{%
\begin{tabular}{lrrrrrrrrrr}
\toprule
 & llama3:70b Standard & llama3:70b Simple & llama3:70b Syntax & llama3:70b Social & llama3:70b Expert & gpt-4 turbo Standard & gpt-4 turbo Simple & gpt-4 turbo Syntax & gpt-4 turbo Social & gpt-4 turbo Expert \\
\midrule
\textbf{llama3:70b Standard} & \textemdash
& \pos{$+0.08$} & \negc{$-0.02$} & \negc{$-0.12$} & \negc{$-0.06$} & \negc{$\boldsymbol{-0.44^{*}}$} & \negc{$-0.22$} & \negc{$\boldsymbol{-0.44^{*}}$} & \negc{$\boldsymbol{-0.46^{*}}$} & \negc{$\boldsymbol{-0.46^{*}}$} \\
\textbf{llama3:70b Simple} & \negc{$-0.08$}
& \textemdash & \negc{$-0.10$} & \negc{$-0.20$} & \negc{$-0.14$} & \negc{$\boldsymbol{-0.52^{*}}$} & \negc{$\boldsymbol{-0.30^{*}}$} & \negc{$\boldsymbol{-0.52^{*}}$} & \negc{$\boldsymbol{-0.54^{*}}$} & \negc{$\boldsymbol{-0.54^{*}}$} \\
\textbf{llama3:70b Syntax} & \pos{$+0.02$}
& \pos{$+0.10$} & \textemdash & \negc{$-0.10$} & \negc{$-0.04$} & \negc{$\boldsymbol{-0.42^{*}}$} & \negc{$-0.20$} & \negc{$\boldsymbol{-0.42^{*}}$} & \negc{$\boldsymbol{-0.44^{*}}$} & \negc{$\boldsymbol{-0.44^{*}}$} \\
\textbf{llama3:70b Social} & \pos{$+0.12$}
& \pos{$+0.20$} & \pos{$+0.10$} & \textemdash & \pos{$+0.06$} & \negc{$\boldsymbol{-0.32^{*}}$} & \negc{$-0.10$} & \negc{$\boldsymbol{-0.32^{*}}$} & \negc{$\boldsymbol{-0.34^{*}}$} & \negc{$\boldsymbol{-0.34^{*}}$} \\
\textbf{llama3:70b Expert} & \pos{$+0.06$}
& \pos{$+0.14$} & \pos{$+0.04$} & \negc{$-0.06$} & \textemdash & \negc{$\boldsymbol{-0.38^{*}}$} & \negc{$-0.16$} & \negc{$\boldsymbol{-0.38^{*}}$} & \negc{$\boldsymbol{-0.40^{*}}$} & \negc{$\boldsymbol{-0.40^{*}}$} \\
\textbf{gpt-4 turbo Standard} & \pos{$\boldsymbol{+0.44^{*}}$}
& \pos{$\boldsymbol{+0.52^{*}}$} & \pos{$\boldsymbol{+0.42^{*}}$} & \pos{$\boldsymbol{+0.32^{*}}$} & \pos{$\boldsymbol{+0.38^{*}}$} & \textemdash & \pos{$+0.22$} & 0.0 & \negc{$-0.02$} & \negc{$-0.02$} \\
\textbf{gpt-4 turbo Simple} & \pos{$+0.22$}
& \pos{$\boldsymbol{+0.30^{*}}$} & \pos{$+0.20$} & \pos{$+0.10$} & \pos{$+0.16$} & \negc{$-0.22$} & \textemdash & \negc{$-0.22$} & \negc{$\boldsymbol{-0.24^{*}}$} & \negc{$\boldsymbol{-0.24^{*}}$} \\
\textbf{gpt-4 turbo Syntax} & \pos{$\boldsymbol{+0.44^{*}}$}
& \pos{$\boldsymbol{+0.52^{*}}$} & \pos{$\boldsymbol{+0.42^{*}}$} & \pos{$\boldsymbol{+0.32^{*}}$} & \pos{$\boldsymbol{+0.38^{*}}$} & 0.0 & \pos{$+0.22$} & \textemdash & \negc{$-0.02$} & \negc{$-0.02$} \\
\textbf{gpt-4 turbo Social} & \pos{$\boldsymbol{+0.46^{*}}$}
& \pos{$\boldsymbol{+0.54^{*}}$} & \pos{$\boldsymbol{+0.44^{*}}$} & \pos{$\boldsymbol{+0.34^{*}}$} & \pos{$\boldsymbol{+0.40^{*}}$} & \pos{$+0.02$} & \pos{$\boldsymbol{+0.24^{*}}$} & \pos{$+0.02$} & \textemdash & 0.00 \\
\textbf{gpt-4 turbo Expert} & \pos{$\boldsymbol{+0.46^{*}}$}
& \pos{$\boldsymbol{+0.54^{*}}$} & \pos{$\boldsymbol{+0.44^{*}}$} & \pos{$\boldsymbol{+0.34^{*}}$} & \pos{$\boldsymbol{+0.40^{*}}$} & \pos{$+0.02$} & \pos{$\boldsymbol{+0.24^{*}}$} & \pos{$+0.02$} & \negc{$-0.00$} & \textemdash \\
\bottomrule
\end{tabular}%
}
\end{table}

\section{User Interface}
\label{gui}

\begin{figure}
  \centering
  \includegraphics[width=0.6\textwidth]{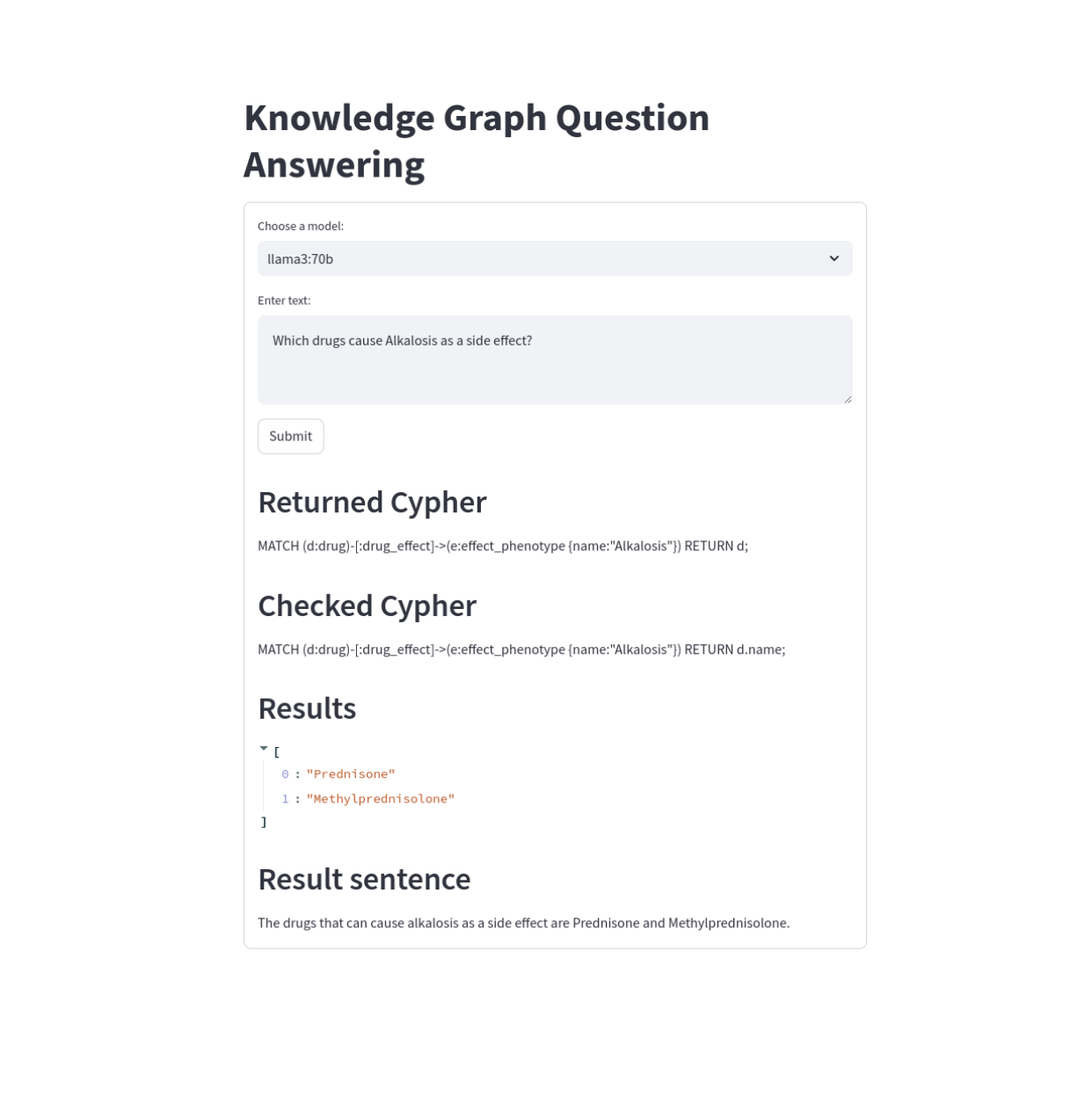}
  \caption{Screenshot of GUI}
  \label{gui_screenshot}
\end{figure}

We created a web-based user interface (UI) for this system specifically designed for interacting with the Knowledge Graph described in \autoref{data} through natural language queries. A screenshot is shown in \autoref{gui_screenshot}. Using this UI, users can input questions, view the initial Cypher query generated by the LLM, and inspect the corrected query after validation by the query-checking algorithm, all within the same interface. This feature allows users to clearly compare the original and corrected queries and check which node and relationship types are being used in the query, providing transparency into the modifications made.

The GUI was developed using Streamlit \cite{streamlit}. It interfaces with Neo4j \cite{neo4j} and Ollama \cite{ollama}, making all downloaded LLMs and a Knowledge Graph available for question answering.

\section{Related Work}
\label{related_work}
Recent advancements in question answering over Knowledge Graphs (KGs) increasingly leverage large language models (LLMs) to enhance query generation and answer formulation. Approaches integrating LLMs with KGs include direct translation methods, prompt optimization, subgraph extraction, and entity matching. This section reviews these methods, highlighting their contributions to KG-based question answering. While we briefly compare non-LLM KGQA methods to contextualize the space, our experimental scope is limited to LLM-based KG querying and prompting under a shared KG and pipeline.

\paragraph{Direct Translation and Fine-Tuning of Language Models}
Several studies focus on directly translating natural language questions into SPARQL queries through fine-tuned LLMs. Wang et al. \cite{wang2023nlqxform} introduced NLQxform, where a BART model is fine-tuned on SPARQL, avoiding ID hallucinations by using entity names rather than IDs and employing string matching for entity recognition. They construct a "logical form" as a query template based on the question, selecting results via template matching. Similarly, Luo et al. \cite{luo2023chatkbqa} proposed ChatKBQA, fine-tuning an LLM on pairs of questions and logical forms of SPARQL queries and generating queries by identifying the top-k similar entities and relations for each candidate form. Fine-tuning LLMs has proven effective for SPARQL generation, minimizing ID hallucinations and improving entity recognition through template matching.

\paragraph{Chain-of-Thought Prompting and Few-Shot Learning}

Chain-of-thought (CoT) prompting enhances LLMs' reasoning in query generation. Zahera et al. \cite{zahera2024generating} utilized CoT prompting and provided extracted entities and relations from the input question, along with few-shot examples that match the question's structure. Their method leverages entity-linking and relation extraction libraries to inform the LLM. Avila et al. \cite{avila2024framework} match question-query pairs to the input question by similarity in question structure. They get passed to the LLM alongside entities and relations and the input question to perform few-shot prompting. Overall, CoT prompting and few-shot learning have shown to enhance the reasoning capabilities of LLMs.

\paragraph{Semantic Parsing and Template-Based Methods}
Semantic parsing has been an important technique in transforming natural language into machine-understandable queries. Pérez et al. \cite{perez2023semantic} explored semantic parsing for conversational question answering with two distinct approaches. The first involved creating a subgraph from identified KG entities and their one-hop neighborhoods, which, along with the question and context, was input into a semantic parser to generate SPARQL queries. The second approach used an encoder-decoder model to convert questions into logical form templates, predicting query structures and relations. Entity recognition and linking were facilitated by a dedicated model, and a graph attention network encoded the graph schema to fill in missing template parts. Agarwal et al. \cite{agarwal2024bring} presented a self-supervised program synthesis approach for zero-shot KGQA. They generated question-query pairs by traversing the KG and having the LLM formulate questions for the paths found. For real questions, similar pairs were retrieved based on cosine similarity of sentence embeddings to serve as few-shot examples. Candidates were re-ranked by regenerating questions from queries to find the closest match. In essence, semantic parsing techniques and template-based methods offer a structured approach for question-answering.

\paragraph{Subgraph Extraction and Contextualization}

Efficient subgraph extraction can help to reduce token usage and focus the LLM on relevant information. Avila et al. \cite{avila2024framework} proposed a framework that retrieves a subgraph centered around the user's question. They grouped triples by subject and property to form sentences, computed embeddings to identify those closest to the question, and reconstructed a connected subgraph. The LLM then generated multiple SPARQL queries, selected the best one based on the results, and formulated a natural language answer. Pliukhin and Rehm's \cite{pliukhin2023improving} method also involved subgraph extraction by deriving two-hop paths containing relevant objects, which were merged and used to prompt the LLM. They also passed a question-query pair that is similar to the question and validated the query by removing hallucinated symbols. Soman et al. \cite{soman2024biomedical} focused on biomedical KGs by extracting disease entities using an LLM and computing vector similarities with disease concept embeddings in SPOKE \cite{morris2023scalable}. They fetched neighboring nodes, converted triples into sentences, and pruned the context by selecting sentences with high similarity scores to the input prompt. This pruned context, along with the question, was provided to the LLM. Pérez et al. \cite{perez2023semantic} constructed a subgraph from KG entities identified in the question and their one-hop neighborhoods. Thus, subgraph extraction techniques that focus on constructing relevant subgraphs from KG entities allow LLMs to generate more accurate queries by providing concise and context-rich input.

\paragraph{Entity and Relation Matching Techniques}

Accurate entity and relation matching is important for the validity of generated queries. Zahera et al. \cite{zahera2024generating} and Wang et al. \cite{wang2023nlqxform} utilized entity-linking and relation extraction libraries with string matching techniques,  respectively, to identify relevant KG components. Steinigen et al. \cite{steinigen2024fact} improved upon this by performing entity extraction to replace synonyms with the corresponding graph entities. Their work, Fact Finder, closely aligns with our approach by incorporating KG schema and relation descriptions into the prompt for the LLM to generate Cypher queries. They also implemented query preprocessing steps such as formatting, lowercasing properties, synonym mapping, and fixing deprecated code issues. Jia et al. \cite{jia2024leveraging} leveraged LLMs for semantic query processing in a scholarly knowledge graph. Their approach involves the LLM generating a triple structure that could potentially answer the user's question. All triples from the KG that fulfill this structure are then extracted. The LLM evaluates these triples to determine which ones answer the question effectively. Entity matching is performed by checking if lowercased labels of entities are contained within the input text, and relevance is assessed based on the frequency of entity labels appearing in clusters of triples. If no exact matches are found, they employ query relaxation by removing parts of the triple or substituting relations with similar ones. Luo et al. \cite{luo2023chatkbqa} enhanced entity and relation matching by calculating similarities for entities and relations in the logical forms and retrieving the top matches. In conclusion, entity and relation matching methods enhance query accuracy by leveraging string matching, synonym mapping, and semantic processing to align input questions with the corresponding graph entities and relations.

\section{Conclusion} 
\label{conclusion}
This research aimed to improve the accuracy and reliability of natural language question-answering systems by integrating Large Language Models (LLMs) with Knowledge Graphs, particularly in the biomedical domain. We developed a pipeline where LLMs generate Cypher queries, which are then validated and corrected using a query-checking algorithm. To evaluate our approach, we created a benchmark dataset of 50 biomedical questions and compared the performance of various LLMs, including GPT-4 Turbo, GPT-5 and Llama 3.3:70b.

Our results show that GPT-4 Turbo and GPT-5 consistently outperformed current open-source models in generating accurate Cypher queries. Although switching from zero-shot prompting to few-shot prompting notably improved the performance of the open-source model Llama 3:70b, it did not reach the accuracy and reliability levels achieved by GPT-4 Turbo. Promptcrafting also did not increase the performance of Llama 3:70b. These findings suggest that while prompt engineering can enhance open-source model performance, a gap remains between open-source and proprietary models in Cypher query generation. When analyzing whether paraphrasing the questions influences performance, we could find no difference for the models we used in the analysis, GPT-4 Turbo and Llama 3.3:70b.

This study also contributes a valuable dataset of 50 biomedical questions and answers tailored to a subset of PrimeKG, offering a resource for future benchmarking and inspiring similar datasets across other Knowledge Graphs.

\section{Outlook} 
\label{outlook}
We identify several avenues to refine the pipeline. Expanding the question set may be informative but is not required for the present conclusions, which target schema-grounded NL-to-Cypher synthesis on 1- to 3-hop queries. A separate research track is entity linking and semantic matching. This targets surface-form resolution rather than query reasoning and can be studied independently on top of our system.

Future work includes integrating a second language model to assist with query validation or refinement. Longer-term, linking individual patient data to the knowledge graph could enable patient-specific querying. This would require data governance, consent, privacy-preserving processing, and harmonization across modalities (e.g., omics, labs, EKG), and sits beyond the present study’s scope.

\vspace{6pt} 

\section*{Funding}
This work was supported by the Deutsche Forschungsgemeinschaft (DFG) (project grant 460135501).

\section*{Data availability}
The source code for generating the results of this paper and for the user-interface can be found in our Git repository: https://git.zib.de/lpusch/cyphergenkg-gui

\section*{Conflicts of interest}
The authors declare no conflicts of interest.

\appendix

\section{PrimeKG structure and changes} 

\label{primekg_changes}

In optimizing PrimeKG for efficient testing, several modifications were made to its structure and names. The graph was first reduced to a 2-hop subgraph centered around 'multiple sclerosis,' cutting it down from around 8 million triples, 30 relational types, and 130,000 entities to about 45,000 triples, 22 unique relational types, and 7,413 distinct entities. This reduction aimed to make queries faster.

Additionally, the graph's relational directionality was altered to make only self-relations bidirectional. This helped streamline information and improve logical consistency in sentence generation.

\begin{tabularx}{\textwidth}{@{}|Y|Y|Y|Y|@{}}   
    \caption{Adjustments Made to PrimeKG}
    \label{primekg_table}\\
    \hline
    \textbf{Type of Change} & \textbf{PrimeKG} & \textbf{Adjusted Version} & \textbf{Reason} \\
    \hline
    \endfirsthead
    \hline
    \textbf{Type of Change} & \textbf{PrimeKG} & \textbf{Adjusted Version} & \textbf{Reason} \\
    \hline
    \endhead
    \hline
    \multicolumn{4}{r}{\textit{Continued on next page}} \\
    \hline
    \endfoot
    \hline
    \endlastfoot            
        2-hop subgraph around 'multiple sclerosis' & $\sim$8M triples, 30 distinct relations, $\sim$130k unique items & $\sim$45k triples, 22 distinct relations, $\sim$7,400 unique items & Make querying graph faster \\
        Relation direction & All relations unidirectional & Only self-relations bidirectional & Shrink graph information, makes more sense in a sentence \\
        \bottomrule
\end{tabularx}

\section{Questions and answers}
\label{question_appendix}

This table contains the 50 questions and answers that were used as a benchmark.

\begin{tabularx}{\textwidth}{@{}|Y|Y|@{}}   
\caption{50 questions}
\label{question_table}\\
\hline
\textbf{Question} & \textbf{Answer}  \\
\hline
\endfirsthead
\hline
\textbf{Question} & \textbf{Answer}\\
\hline
\endhead
\hline
\multicolumn{2}{r}{\textit{Continued on next page}} \\
\hline
\endfoot
\hline
\endlastfoot
What are the names of the drugs that are contraindicated when a patient has multiple sclerosis? & Ascorbic acid, Zinc gluconate\\
Which drugs are contraindicated when I have dermatitis? & Hydrocortisone, Cortisone acetate, Triamcinolone\\
Which drugs cause Alkalosis as a side effect? & Methylprednisolone, Prednisone\\
Which drugs have Anxiety as side effect? & Methylprednisolone, Prednisone, Hydrocortisone, Dexamethasone, Betamethasone\\
Which genes are expressed in the eye? & CBLB, RBPJ, KCNJ10, SLC11A1, CLEC16A\\
What genes are expressed in the nasopharynx? & RBPJ\\
What proteins interact with GRB2? & CBLB, GC, IL7R, P2RX7, TNFRSF1A, VCAM1\\
What proteins interact with the protein KPNA2? & CBLB, IL7R, IL10, VCAM1\\
What are off-label uses of Zinc gluconate? & methemoglobinemia, sulfhemoglobinemia, attention deficit-hyperactivity disorder, attention deficit hyperactivity disorder, inattentive type, gastroenteritis, acrodermatitis enteropathica\\
What are off-label uses for Ascorbic acid? & Coronavinae infectious disease\\
In which cellular components is a protein expressed that is associated with the Spasticity phenotype? & plasma membrane, cytoplasm, integral component of plasma membrane, membrane raft, actin cytoskeleton, glutamatergic synapse, mitochondrial outer membrane, integral component of presynaptic membrane, GABA-ergic synapse, growth cone, integral component of mitochondrial membrane\\
In which cellular components is a protein expressed that is associated with Nausea? & cytoplasm, extracellular space, extracellular region, secretory granule, secretory granule lumen\\
What side effects does a drug have that is indicated for Richter syndrome? & Abdominal distention, Abdominal pain, Adrenal insufficiency, Alkalosis, Alopecia, Alopecia of scalp, Anaphylactic shock, Poor appetite, Anxiety, Arrhythmia, Arthralgia, Hypertrophic cardiomyopathy, Corneal ulceration, Delusions, Inflammatory abnormality of the skin, Atopic dermatitis, Vertigo, Bruising susceptibility, Edema, Abnormality of the endocrine system, Seizure, Abnormality of the eye, Fatigue, Fever, Erythema, Recurrent fractures, Gastrointestinal hemorrhage, Glycosuria, Hallucinations, Headache, Cardiac arrest, Cardiomegaly, Congestive heart failure, Hepatomegaly, Hirsutism, Hypercholesterolemia, Hyperglycemia, Hypernatremia, Hyperthyroidism, Hypothyroidism, Abnormal joint morphology, Arthropathy, Lethargy, Leukocytosis, Nausea, Abnormal peripheral nervous system morphology, Polyneuropathy, Peripheral neuropathy, Avascular necrosis, Osteoporosis, Generalized osteoporosis, Pancreatitis, Optic neuritis, Papilledema, Paraplegia, Paresthesia, Peptic ulcer, Petechiae, Pruritus, Pulmonary edema, Facial erythema, Loss of consciousness, Syncope, Tachycardia, Telangiectasia, Thrombophlebitis, Vasculitis, Vomiting, Increased body weight, Agitation, Emotional lability, Mood swings, Mood changes, Dermal atrophy, EEG abnormality, Impaired glucose tolerance, Growth delay, Increased intracranial pressure, Muscle weakness, Tendon rupture, Striae distensae, Irregular menstruation, Malnutrition, Abnormality of the skin, Paraparesis, Myalgia, Polyphagia, Memory impairment, Ocular hypertension, Subcapsular cataract, Personality changes, Vertebral compression fractures, Myopathy, Lipoatrophy, Mania, Blurred vision, Scaling skin, Hyperactivity, Hyperkinetic movements, Bradycardia, Dementia, Facial edema, Hyperhidrosis, Dry skin\\
If I take drugs for dry eye syndrome, what side effects will they have? & Abdominal distention, Anaphylactic shock, Arrhythmia, Hypertrophic cardiomyopathy, Corneal ulceration, Atopic dermatitis, Bruising susceptibility, Edema, Inflammatory abnormality of the skin, Headache, Cardiac arrest, Cardiomegaly, Congestive heart failure, Hepatomegaly, Hirsutism, Hypernatremia, Abnormal joint morphology, Arthropathy, Keratitis, Mydriasis, Nausea, Abnormal peripheral nervous system morphology, Polyneuropathy, Peripheral neuropathy, Osteoporosis, Generalized osteoporosis, Pancreatitis, Optic neuritis, Papilledema, Paraplegia, Paresthesia, Peptic ulcer, Petechiae, Pulmonary edema, Seizure, Vertigo, Loss of consciousness, Syncope, Tachycardia, Thrombophlebitis, Vasculitis, Increased body weight, Hypokalemic alkalosis, Emotional lability, Mood swings, Mood changes, Avascular necrosis, Increased intracranial pressure, Muscle weakness, Tendon rupture, Striae distensae, Irregular menstruation, Paraparesis, Polyphagia, Ocular hypertension, Subcapsular cataract, Erythema, Personality changes, Vertebral compression fractures, Myopathy, Blurred vision, Bradycardia, Visual impairment, Pain, Hyperhidrosis, Dry skin\\
In what anatomical structures is there no expression of proteins that interact with leukocyte migration? & cerebellar vermis\\
In what anatomical structures is there no expression of proteins that interact with cell migration? & vastus lateralis, cerebellar vermis\\
What genes and biological processes does an exposure to Tobacco Smoke Pollution interact with? & regulation of blood pressure, triglyceride metabolic process, respiratory system process, gene expression, DNA methylation, spermatogenesis, cognition, regulation of DNA methylation, developmental growth, immune response, cholesterol metabolic process, behavior, lipid metabolic process, DNA methylation on cytosine within a CG sequence, inflammatory response, regulation of gene silencing by miRNA, regulation of respiratory gaseous exchange, DNA metabolic process, respiratory gaseous exchange by respiratory system, menopause, circulatory system process, mRNA methylation, feeding behavior, hypersensitivity, estrone secretion, alanine metabolic process, lactate metabolic process, IFNG, IL1B, ARNT, ATF6B, BNIP3L, DDB2, FTH1, GADD45A, RAD51, TP53, TXN, AHRR, CNTNAP2, CYP1A1, EXT1, GFI1, HLA-DPB2, MYO1G, RUNX1, TTC7B, F2RL3, SLC7A8, C11orf52, FRMD4A, IL1B, IFNG, IL4, TNF, SERPINE1\\
What genes and biological processes does an exposure to Lead interact with? & regulation of blood pressure, gene expression, cognition, head development, regulation of DNA methylation, glucose metabolic process, mitochondrial DNA metabolic process, behavior, lipid metabolic process, DNA methylation on cytosine within a CG sequence, regulation of heart rate, memory, regulation of systemic arterial blood pressure, response to oxidative stress, psychomotor behavior, hemoglobin biosynthetic process, visual perception, developmental process involved in reproduction, metabolic process, regulation of humoral immune response mediated by circulating immunoglobulin, glomerular filtration, cortisol metabolic process, detection of oxidative stress, tissue homeostasis, social behavior, calcium ion homeostasis, heart contraction, humoral immune response, lymphocyte mediated immunity, transport, ethanolamine metabolic process, glutamate metabolic process, urea metabolic process, regulation of cortisol secretion, inositol metabolic process, DNA methylation involved in gamete generation, regulation of multicellular organism growth, regulation of amyloid-beta formation, regulation of genetic imprinting, positive regulation of multicellular organism growth, sensory perception of sound, homocysteine metabolic process, response to auditory stimulus, renal filtration, response to lead ion, detection of mechanical stimulus involved in sensory perception, cellular amine metabolic process, choline metabolic process, creatine metabolic process, brain development, ICAM1, ADAM9, LRPAP1, RTN4, APP, IL6, TNFRSF1B, CRP, ICAM1, H19, HYMAI, IGF2, PEG3, PLAGL1, MIR10A, MIR146A, MIR190B, MIR431, MIR651, IGF1, HEXB, B2M, MIR222, ALB, PON1\\
With which pathways do proteins interact that are associated with sleep-wake disorder? & Interleukin-1 processing, Pyroptosis, CLEC7A/inflammasome pathway, Interleukin-10 signaling, Interleukin-4 and Interleukin-13 signaling, Interleukin-1 signaling, Purinergic signaling in leishmaniasis infection, Opioid Signalling, Androgen biosynthesis, Glucocorticoid biosynthesis, G-protein activation, Peptide hormone biosynthesis, Endogenous sterols, Peptide ligand-binding receptors, G alpha (s) signalling events, G alpha (i) signalling events, Defective ACTH causes obesity and POMCD, FOXO-mediated transcription of oxidative stress, metabolic and neuronal genes, ADORA2B mediated anti-inflammatory cytokines production\\
With which pathways do proteins interact that are associated with sickle cell anemia? & Immunoregulatory interactions between a Lymphoid and a non-Lymphoid cell, Integrin cell surface interactions, Interleukin-4 and Interleukin-13 signaling, Interferon gamma signaling\\
What are phenotypes that gene POMC is associated with that also occur in neuromyelitis optica? & Ocular pain, Nausea\\
What are phenotypes that gene IFNG is associated with that also occur in neuromyelitis optica? & Nausea\\
What drugs should I take if I have a disease because of an exposure to Lead? & Methylprednisolone, Prednisone, Dalfampridine, Prednisolone, Hydrocortisone, Cortisone acetate, Hydrocortisone acetate, Dexamethasone, Betamethasone, Natalizumab, Teriflunomide, Ozanimod, Triamcinolone\\
What drugs should I take if I have a disease because of an exposure to Tobacco Smoke Pollution? & Methylprednisolone, Prednisone, Dalfampridine, Prednisolone, Hydrocortisone, Cortisone acetate, Hydrocortisone acetate, Dexamethasone, Betamethasone, Natalizumab, Teriflunomide, Ozanimod, Triamcinolone\\
What diseases are the diseases where Dalfampridine is contraindicated for related to? & brain disease\\
What diseases are the diseases where Morphine is contraindicated for related to? & multiple sclerosis, megalencephalic leukoencephalopathy with cysts, encephalopathy, acute, infection-induced, diabetic encephalopathy, hydrocephalus, brain compression, cerebral sarcoidosis, hepatic encephalopathy, visual pathway disease, central nervous system origin vertigo, cerebellar disease, olfactory nerve disease, thalamic disease, pituitary gland disease, disorder of optic chiasm, basal ganglia disease, epilepsy, mental disorder, subarachnoid hemorrhage (disease), central nervous system cyst (disease), migraine disorder, prion disease, delayed encephalopathy after acute carbon monoxide poisoning, cerebral malaria, akinetic mutism, Reye syndrome, brain edema, encephalomalacia, intracranial hypertension, intracranial hypotension, kernicterus, Wernicke encephalopathy, encephalopathy, recurrent, of childhood, progressive bulbar palsy, cerebrovascular disorder, disorder of medulla oblongata, brain inflammatory disease, narcolepsy-cataplexy syndrome, meningoencephalocele, cerebral sinovenous thrombosis, autoimmune encephalopathy with parasomnia and obstructive sleep apnea, neurometabolic disease, cerebral organic aciduria, narcolepsy without cataplexy, cerebral lipidosis with dementia, brain neoplasm, colpocephaly, corpus callosum agenesis of blepharophimosis robin type, corpus callosum dysgenesis X-linked recessive, corpus callosum dysgenesis cleft spasm, corpus callosum dysgenesis hypopituitarism, cerebral degeneration, brain injury, encephalopathy, cluster headache syndrome, cerebral cortex disease, midbrain disease, central nervous system disease\\
Which cellular components do the proteins an exposure to Lead affects interact with? & extracellular space, extracellular exosome, collagen-containing extracellular matrix, cell surface, plasma membrane, membrane, integral component of plasma membrane, external side of plasma membrane, membrane raft, focal adhesion, immunological synapse\\
Which cellular components do the proteins an exposure to Tobacco Smoke Pollution affects interact with? & extracellular space, extracellular region, cytosol, lysosome\\
What genes are associated with diseases that are linked to an exposure to Lead? & APOE, BCHE, CASP1, CBLB, CD6, CD40, CD58, CNR1, GC, HLA-DPB1, HLA-DQB1, HLA-DRA, HLA-DRB1, ICAM1, IRF8, IFNB1, IFNG, RBPJ, IL1B, IL1RN, IL2RA, IL7, IL7R, IL10, IL12A, IL17A, KCNJ10, MCAM, CLDN11, P2RX7, PDCD1, POMC, NECTIN2, SELE, SLC11A1, STAT4, TNFAIP3, TNFRSF1A, TYK2, VCAM1, VDR, TNFSF14, KIF1B, CLEC16A, NLRP3\\
What genes are associated with diseases that are linked to an exposure to Mercury? & APOE, BCHE, CASP1, CBLB, CD6, CD40, CD58, CNR1, GC, HLA-DPB1, HLA-DQB1, HLA-DRA, HLA-DRB1, ICAM1, IRF8, IFNB1, IFNG, RBPJ, IL1B, IL1RN, IL2RA, IL7, IL7R, IL10, IL12A, IL17A, KCNJ10, MCAM, CLDN11, P2RX7, PDCD1, POMC, NECTIN2, SELE, SLC11A1, STAT4, TNFAIP3, TNFRSF1A, TYK2, VCAM1, VDR, TNFSF14, KIF1B, CLEC16A, NLRP3\\
What side effects of the drug Methylprednisolone are similar to the multiple sclerosis phenotype? & Emotional lability, Paresthesia, Muscle Weakness, Paraplegia, Optic neuritis, Nausea\\
What side effects of the drug Prednisone are similar to the multiple sclerosis phenotype? & Paresthesia, Optic neuritis, Muscle weakness, Emotional lability, Nausea, Paraplegia\\
Which drug is contraindicated in a disease that was linked to an exposure to something that interacts with the protein IFNG? & Ascorbic acid, Zinc gluconate, Methylprednisolone, Prednisone, Prednisolone, Hydrocortisone, Cortisone acetate, Hydrocortisone acetate, Dexamethasone, Betamethasone, Triamcinolone\\
Which drug is contraindicated in a disease that was linked to an exposure to something that interacts with the protein IL1B? & Ascorbic acid, Zinc gluconate, Methylprednisolone, Prednisone, Prednisolone, Hydrocortisone, Cortisone acetate, Hydrocortisone acetate, Dexamethasone, Betamethasone, Triamcinolone\\
What pathways do the exposures that can lead to multiple sclerosis interact with? & Immunoregulatory interactions between a Lymphoid and a non-Lymphoid cell, Integrin cell surface interactions, Interleukin-10 signaling, Interleukin-4 and Interleukin-13 signaling, Interferon gamma signaling, Regulation of IFNG signaling, RUNX1 and FOXP3 control the development of regulatory T lymphocytes (Tregs), Gene and protein expression by JAK-STAT signaling after Interleukin-12 stimulation, Interleukin-1 processing, Pyroptosis, CLEC7A/inflammasome pathway, Interleukin-1 signaling, Purinergic signaling in leishmaniasis infection\\
What pathways do the exposures that can lead to atopic eczema interact with? & Immunoregulatory interactions between a Lymphoid and a non-Lymphoid cell, Integrin cell surface interactions, Interleukin-10 signaling, Interleukin-4 and Interleukin-13 signaling, Interferon gamma signaling\\
Which exposure can affect drugs that are approved for off-label-use for dermatitis? & Chlorpyrifos, glyphosate, Insecticides, Organophosphates, Pesticides, Lead, Tobacco Smoke Pollution\\
Which exposure can affect drugs that are approved for off-label-use for heart disease? & Chlorpyrifos, glyphosate, Insecticides, Organophosphates, Pesticides, Lead, Tobacco Smoke Pollution\\
Which drugs have synergistic interactions with drugs that are affected by proteins that CASP1 has protein-protein interactions with? & Methylprednisolone, Prednisone, Prednisolone, Hydrocortisone, Cortisone acetate, Hydrocortisone acetate, Dexamethasone, Zinc gluconate, Betamethasone, Natalizumab, Teriflunomide, Ozanimod, Triamcinolone\\
Which drugs have synergistic interactions with drugs that are affected by proteins that IL1B has protein-protein interactions with? & Methylprednisolone, Prednisone, Dalfampridine, Prednisolone, Hydrocortisone, Cortisone acetate, Hydrocortisone acetate, Dexamethasone, Zinc gluconate, Betamethasone, Triamcinolone\\
Which biological processes are affected by the gene APOE which are also affected by an exposure to something that is linked to multiple sclerosis? & cholesterol homeostasis, triglyceride metabolic process, cholesterol metabolic process, gene expression\\
Which biological processes are affected by the gene IL1B which are also affected by an exposure to something that is linked to multiple sclerosis? & inflammatory response, immune response\\
What drugs should I not take for a disease that I got because exposure to Tobacco Smoke Pollution interacts with a protein relevant to that disease? & Ascorbic acid, Zinc gluconate, Methylprednisolone, Prednisone, Dalfampridine, Prednisolone, Hydrocortisone, Cortisone acetate, Hydrocortisone acetate, Dexamethasone, Betamethasone, Ozanimod, Triamcinolone, Tolvaptan, Nelarabine\\
What drugs should I not take for a disease that I got because exposure to Lead interacts with a protein relevant to that disease? & Ascorbic acid, Zinc gluconate, Methylprednisolone, Prednisone, Prednisolone, Hydrocortisone, Cortisone acetate, Hydrocortisone acetate, Dexamethasone, Betamethasone, Ozanimod, Triamcinolone, Tolvaptan, Nelarabine\\
What side effects does Prednisone have that also occur when a protein is expressed that is influenced by exposure to Tobacco Smoke Pollution? & Memory impairment, Fever, Leukocytosis, Lethargy, Cardiomegaly, Nausea, Seizure, Vomiting\\
What side effects does Dexamethasone have that also occur when a protein is expressed that is influenced by exposure to Tobacco Smoke Pollution? & Fever, Leukocytosis, Lethargy, Cardiomegaly, Nausea, Seizure, Vomiting\\
What drugs can I take that are indicated for a disease whose phenotype is associated with the gene POMC? & Eculizumab\\
What drugs can I take that are indicated for a disease whose phenotype is associated with the gene IFNG? & Eculizumab\\
What drugs can I take that are approved for off-label-use for a disease that I got because exposure to Tobacco Smoke Pollution interacts with a protein relevant to that disease? & Methylprednisolone, Prednisone, Prednisolone, Hydrocortisone, Cortisone acetate, Dexamethasone, Betamethasone, Triamcinolone\\
What drugs can I take that are approved for off-label-use for a disease that I got because exposure to Particulate Matter interacts with a protein relevant to that disease? & Methylprednisolone, Prednisone, Prednisolone, Hydrocortisone, Cortisone acetate, Dexamethasone, Betamethasone, Triamcinolone\\
\bottomrule
\end{tabularx}

\section{Paraphrased Questions}
\label{sec:paraphrased}
\begin{longtable}{|p{0.48\textwidth} p{0.48\textwidth}|}
\caption{Original questions and paraphrases}\\
\hline
\textbf{Question} & \textbf{Paraphrase} \\
\hline
\endfirsthead
\hline
\textbf{Question} & \textbf{Paraphrase} \\
\hline
\endhead
What are the names of the drugs that are contraindicated when a patient has multiple sclerosis? & Which medications are contraindicated for patients with multiple sclerosis? \\
Which drugs are contraindicated when I have dermatitis? & Which medications are contraindicated for patients with dermatitis? \\
Which drugs cause Alkalosis as a side effect? & Which medications list Alkalosis as an adverse effect? \\
Which drugs have Anxiety as side effect? & Which medications list Anxiety as a side effect? \\
Which genes are expressed in the eye? & Which genes show expression in the eye? \\
What genes are expressed in the nasopharynx? & Which genes show expression in the nasopharynx? \\
What proteins interact with GRB2? & Which proteins interact with GRB2? \\
What proteins interact with the protein KPNA2? & Which proteins interact with KPNA2? \\
What are off-label uses of Zinc gluconate? & Which off-label indications are reported for Zinc gluconate? \\
What are off-label uses for Ascorbic acid? & Which off-label indications are reported for Ascorbic acid? \\
In which cellular components is a protein expressed that is associated with the Spasticity phenotype? & Which cellular components show expression of the protein associated with Spasticity? \\
In which cellular components is a protein expressed that is associated with Nausea? & Which cellular components show expression of the protein associated with Nausea? \\
What side effects does a drug have that is indicated for Richter syndrome? & What adverse effects are associated with a drug indicated for Richter syndrome? \\
If I take drugs for dry eye syndrome, what side effects will they have? & What adverse effects occur with medications used to treat dry eye syndrome? \\
In what anatomical structures is there no expression of proteins that interact with leukocyte migration? & Which anatomical structures lack expression of proteins that interact with leukocyte migration? \\
In what anatomical structures is there no expression of proteins that interact with cell migration? & Which anatomical structures lack expression of proteins that interact with cell migration? \\
What genes and biological processes does an exposure to Tobacco Smoke Pollution interact with? & Which genes and biological processes interact with exposure to Tobacco Smoke Pollution? \\
What genes and biological processes does an exposure to Lead interact with? & Which genes and biological processes interact with Lead exposure? \\
With which pathways do proteins interact that are associated with sleep-wake disorder? & Which pathways do proteins associated with sleep-wake disorder interact with? \\
With which pathways do proteins interact that are associated with sickle cell anemia? & Which pathways do proteins associated with sickle cell anemia interact with? \\
What are phenotypes that gene POMC is associated with that also occur in neuromyelitis optica? & Which phenotypes linked to POMC also occur in neuromyelitis optica? \\
What are phenotypes that gene IFNG is associated with that also occur in neuromyelitis optica? & Which phenotypes linked to the gene IFNG also occur in neuromyelitis optica? \\
What drugs should I take if I have a disease because of an exposure to Lead? & Which medications are indicated for a disease caused by Lead exposure? \\
What drugs should I take if I have a disease because of an exposure to Tobacco Smoke Pollution? & Which medications are indicated for a disease caused by Tobacco Smoke Pollution exposure? \\
What diseases are the diseases where Dalfampridine is contraindicated for related to? & Which diseases are related to the diseases for which Dalfampridine is contraindicated? \\
What diseases are the diseases where Morphine is contraindicated for related to? & Which diseases are related to the diseases for which Morphine is contraindicated? \\
Which cellular components do the proteins an exposure to Lead affects interact with? & Which cellular components do proteins affected by Lead exposure interact with? \\
Which cellular components do the proteins an exposure to Tobacco Smoke Pollution affects interact with? & Which cellular components do proteins affected by Tobacco Smoke Pollution exposure interact with? \\
What genes are associated with diseases that are linked to an exposure to Lead? & Which genes are associated with diseases linked to Lead exposure? \\
What genes are associated with diseases that are linked to an exposure to Mercury? & Which genes are associated with diseases linked to Mercury exposure? \\
What side effects of the drug Methylprednisolone are similar to the multiple sclerosis phenotype? & Which adverse effects of Methylprednisolone overlap with multiple sclerosis phenotypes? \\
What side effects of the drug Prednisone are similar to the multiple sclerosis phenotype? & Which adverse effects of Prednisone overlap with multiple sclerosis phenotypes? \\
Which drug is contraindicated in a disease that was linked to an exposure to something that interacts with the protein IFNG? & Which drug should be avoided for a disease linked to an exposure that interacts with the protein IFNG? \\
Which drug is contraindicated in a disease that was linked to an exposure to something that interacts with the protein IL1B? & Which drug should be avoided for a disease linked to an exposure that interacts with IL1B? \\
What pathways do the exposures that can lead to multiple sclerosis interact with? & Which pathways are modulated by exposures associated with multiple sclerosis? \\
What pathways do the exposures that can lead to atopic eczema interact with? & Which pathways are modulated by exposures associated with atopic eczema? \\
Which exposure can affect drugs that are approved for off-label-use for dermatitis? & Which exposure can modify the response to off-label dermatitis medications? \\
Which exposure can affect drugs that are approved for off-label-use for heart disease? & Which exposure can modify the response to off-label heart disease medications? \\
Which drugs have synergistic interactions with drugs that are affected by proteins that CASP1 has protein-protein interactions with? & Which agents show synergy with drugs affected by CASP1-interacting proteins? \\
Which drugs have synergistic interactions with drugs that are affected by proteins that IL1B has protein-protein interactions with? & Which agents show synergy with drugs affected by IL1B-interacting proteins? \\
Which biological processes are affected by the gene APOE which are also affected by an exposure to something that is linked to multiple sclerosis? & Which processes regulated by the gene APOE are likewise affected by exposures associated with multiple sclerosis? \\
Which biological processes are affected by the gene IL1B which are also affected by an exposure to something that is linked to multiple sclerosis? & Which processes regulated by the gene IL1B are likewise affected by exposures associated with multiple sclerosis? \\
What drugs should I not take for a disease that I got because exposure to Tobacco Smoke Pollution interacts with a protein relevant to that disease? & Which medications should I avoid for a disease that arose because exposure to Tobacco Smoke Pollution interacts with a disease-relevant protein? \\
What drugs should I not take for a disease that I got because exposure to Lead interacts with a protein relevant to that disease? & Which medications should I avoid for a disease that arose because exposure to Lead interacts with a disease-relevant protein? \\
What side effects does Prednisone have that also occur when a protein is expressed that is influenced by exposure to Tobacco Smoke Pollution? & Which adverse effects of Prednisone also occur when a phenotype influenced by Tobacco Smoke Pollution is expressed? \\
What side effects does Dexamethasone have that also occur when a protein is expressed that is influenced by exposure to Tobacco Smoke Pollution? & Which adverse effects of Dexamethasone also occur when a phenotype influenced by Tobacco Smoke Pollution is expressed? \\
What drugs can I take that are indicated for a disease whose phenotype is associated with the gene POMC? & Which medications are indicated for a disease whose phenotype is associated with the gene POMC? \\
What drugs can I take that are indicated for a disease whose phenotype is associated with the gene IFNG? & Which medications are indicated for a disease whose phenotype is associated with the gene IFNG? \\
What drugs can I take that are approved for off-label-use for a disease that I got because exposure to Tobacco Smoke Pollution interacts with a protein relevant to that disease? & Which medications approved for off-label use for the disease are suitable when exposure to Tobacco Smoke Pollution interacts with a disease-relevant protein? \\
What drugs can I take that are approved for off-label-use for a disease that I got because exposure to Particulate Matter interacts with a protein relevant to that disease? & Which medications approved for off-label use for the disease are suitable when exposure to Particulate Matter interacts with a disease-relevant protein? \\
\hline
\end{longtable}

\section{Cypher generation prompts} 
\label{prompts}

\subsection{Zero-shot} 
\label{zero_shot}
\begin{verbatim}
Task:Generate Cypher statement to query a graph database.
Instructions:
Use only the provided relationship types and properties in the schema.
Do not use any other relationship types or properties that are not provided.
The cypher statement should only return nodes that are specifically asked for in the question.
Absolutely do not use the asterisk operator (*) in the cypher statement. It is a little star sign next to the relation. Do not use it!
Schema:
{schema}
Note: Do not include any explanations or apologies in your responses.
Do not respond to any questions that might ask anything else than for you to construct a 
Cypher statement.
Do not include any text except the generated Cypher statement.

The question is:
{question}
\end{verbatim}
  
\subsection{One-shot}
\label{one-shot}
\begin{verbatim}
Task:Generate Cypher statement to query a graph database.
Instructions:
Use only the provided relationship types and properties in the schema.
Do not use any other relationship types or properties that are not provided.
The cypher statement should only return nodes that are specifically asked for in the question.
Absolutely do not use the asterisk operator (*) in the cypher statement. It is a little star sign next to the relation. Do not use it!
Schema:
{schema}
Note: Do not include any explanations or apologies in your responses.
Do not respond to any questions that might ask anything else than for you to construct a Cypher statement.
Do not include any text except the generated Cypher statement.
Follow these Cypher example when Generating Cypher statements:
# How many actors played in Top Gun?
MATCH (m:movie {{name:"Top Gun"}})<-[:acted_in]-(a:actor)
RETURN a.name

The question is:
{question}
\end{verbatim}

\subsection{Few-shot} 
\label{few_shot}
\begin{verbatim}
Task:Generate Cypher statement to query a graph database.
Instructions:
Use only the provided relationship types and properties in the schema.
Do not use any other relationship types or properties that are not provided.
The cypher statement should only return nodes that are specifically asked for in the question.
Absolutely do not use the asterisk operator (*) in the cypher statement. It is a little star sign next to the relation. Do not use it!
Schema:
{schema}
Note: Do not include any explanations or apologies in your responses.
Do not respond to any questions that might ask anything else than for you to construct a Cypher statement.
Do not include any text except the generated Cypher statement.
Follow these Cypher example when Generating Cypher statements:
# Which actors played in Top Gun?
MATCH (m:movie {{name:"Top Gun"}})<-[:acted_in]-(a:actor)
RETURN a.name
# What town were the actors that played in Top Gun born in?
MATCH (m:movie {{name:"Top Gun"}})<-[:acted_in]-(a:actor)-[:born_in]->(t:town)
RETURN t.name
# What are the mayors of the towns that the actors that played in Top Gun were born in?
MATCH (m:movie {{name:"Top Gun"}})<-[:acted_in]-(a:actor)-[:born_in]->(t:town)<-[:is_mayor]-(m:mayor)
RETURN m.name

The question is:
{question}
\end{verbatim}

\subsection{Simple prompt} 
\label{simple_prompt}

\begin{verbatim}
Task:Generate Cypher statement to query a graph database.
Schema:
{schema}
Note: Do not include any explanations or apologies in your responses.
Do not respond to any questions that might ask anything else than for you to construct a Cypher statement.
Do not include any text except the generated Cypher statement.

The question is:
{question}
\end{verbatim}

\subsection{Syntax prompt} 
\label{syntax_prompt}
\begin{verbatim}
Task:Generate Cypher statement to query a graph database.
Instructions:
Use only the provided relationship types and properties in the schema.
Do not use any other relationship types or properties that are not provided.
The cypher statement should only return nodes that are specifically asked for in the question.
Absolutely do not use the asterisk operator (*) in the cypher statement. It is a little star sign next to the relation. Do not use it!
Please pay attention only to use valid cypher syntax!
Schema:
{schema}
Note: Do not include any explanations or apologies in your responses.
Do not respond to any questions that might ask anything else than for you to construct a Cypher statement.
Do not include any text except the generated Cypher statement.

The question is:
{question}
\end{verbatim}

\subsection{Social engineering prompt}
\label{social_prompt}
\begin{verbatim}
Task:Generate Cypher statement to query a graph database. It is extremely important that you don't make any mistakes, or I will get fired!
Instructions:
Use only the provided relationship types and properties in the schema.
Do not use any other relationship types or properties that are not provided.
The cypher statement should only return nodes that are specifically asked for in the question.
Absolutely do not use the asterisk operator (*) in the cypher statement. It is a little star sign next to the relation. Do not use it!
Schema:
{schema}
Note: Do not include any explanations or apologies in your responses.
Do not respond to any questions that might ask anything else than for you to construct a Cypher statement.
Do not include any text except the generated Cypher statement.

The question is:
{question}
\end{verbatim}

\subsection{Role prompt} 
\label{role_prompt}
\begin{verbatim}
You are a very knowledgeable cypher query expert with years of experience. 
Task: Generate Cypher statement to query a graph database.
Instructions:
Use only the provided relationship types and properties in the schema.
Do not use any other relationship types or properties that are not provided.
The cypher statement should only return nodes that are specifically asked for in the question.
Absolutely do not use the asterisk operator (*) in the cypher statement. It is a little star sign next to the relation. Do not use it!
Schema:
{schema}
Note: Do not include any explanations or apologies in your responses.
Do not respond to any questions that might ask anything else than for you to construct a Cypher statement.
Do not include any text except the generated Cypher statement.

The question is:
{question}
\end{verbatim}

\bibliographystyle{unsrtnat}
\bibliography{references}

@article{chandak2022building,
  title={Building a knowledge graph to enable precision medicine},
  author={Chandak, Payal and Huang, Kexin and Zitnik, Marinka},
  journal={Nature Scientific Data},
  doi={https://doi.org/10.1038/s41597-023-01960-3},
  URL={https://www.nature.com/articles/s41597-023-01960-3},
  year={2023}
}

@misc{cypher_langchain,
    title        = {Neo4j DB QA chain},
    author       = {LangChain},
    year         = 2025,
    note         = {\url{https://python.langchain.com/docs/tutorials/graph/} [Accessed: 2025-09-11]}
}

@misc{langchain_graph,
author = {{Neo4j}},
title  = {{Neo4jGraph}},
url    = {https://api.python.langchain.com/en/latest/graphs/langchain_community.graphs.neo4j_graph.Neo4jGraph.html},
note   = {Accessed: 2025-09-11}
}

@article{rag,
  title={Retrieval-augmented generation for knowledge-intensive nlp tasks},
  author={Lewis, Patrick and Perez, Ethan and Piktus, Aleksandra and Petroni, Fabio and Karpukhin, Vladimir and Goyal, Naman and K{\"u}ttler, Heinrich and Lewis, Mike and Yih, Wen-tau and Rockt{\"a}schel, Tim and others},
  journal={Advances in Neural Information Processing Systems},
  volume={33},
  pages={9459--9474},
  year={2020}
}

@misc{langchain,
author = {{LangChain}},
title  = {{LangChain}},
url    = {https://github.com/langchain-ai/langchain},
note   = {Accessed: 2025-09-11}
}

@misc{ollama,
author = {{Ollama}},
title  = {{Ollama}},
url    = {https://ollama.com/},
note   = {Accessed: 2025-09-11}
}

@article{xu2024hallucination,
  title={Hallucination is inevitable: An innate limitation of large language models},
  author={Xu, Ziwei and Jain, Sanjay and Kankanhalli, Mohan},
  journal={arXiv preprint arXiv:2401.11817},
  year={2024}
}

@online{openai_introducing_gpt5_2025,
  author    = {OpenAI},
  title     = {Introducing GPT-5},
  year      = {2025},
  month     = aug,
  date      = {2025-08-07},
  url       = {https://openai.com/index/introducing-gpt-5/},
  urldate   = {2025-10-31}
}

@misc{lcel,
author = {{LangChain}},
title  = {{LangChain Expression Language (LCEL)}},
url    = {https://python.langchain.com/v0.1/docs/expression_language/},
note   = {Accessed: 2025-09-11}
}

@online{meta_llama_3_3_70b_instruct_2024,
  author    = {Meta AI},
  title     = {Llama 3.3 70B Instruct - Model Card},
  year      = {2024},
  month     = dec,
  date      = {2024-12-06},
  url       = {https://huggingface.co/meta-llama/Llama-3.3-70B-Instruct},
  urldate   = {2025-10-31},
  note      = {Model card}
}

@misc{streamlit,
author = {{Streamlit}},
title  = {{Streamlit}},
url    = {https://streamlit.io/},
note   = {Accessed: 2025-09-11}
}

@misc{neo4j,
author = {{Neo4j}},
title  = {{Neo4j}},
url    = {https://neo4j.com},
note   = {Accessed: 2025-09-11}
}

@inproceedings{avila2024framework,
  title={A Framework for Question Answering on Knowledge Graphs Using Large Language Models},
  author={Avila, Caio Viktor S and Casanova, Marco A and Vidal, V{\^a}nia MP},
  year={2024},
  organization={ESWC}
}

@misc{zahera2024generating,
  title={Generating sparql from natural language using chain-of-thoughts prompting},
  author={Zahera, Hamada M and Ali, Manzoor and Sherif, Mohamed Ahmed and Moussallem, Diego and Ngomo, A-C Ngonga},
  year={2024},
  publisher={SEMANTiCS}
}

@inproceedings{pliukhin2023improving,
  title={Improving Subgraph Extraction Algorithms for One-Shot SPARQL Query Generation with Large Language Models.},
  author={Pliukhin, Dmitrii and Radyush, Daniil and Kovriguina, Liubov and Mouromtsev, Dmitry},
  booktitle={QALD/SemREC@ ISWC},
  year={2023}
}

@article{soman2024biomedical,
  title={Biomedical knowledge graph-optimized prompt generation for large language models},
  author={Soman, Karthik and Rose, Peter W and Morris, John H and Akbas, Rabia E and Smith, Brett and Peetoom, Braian and Villouta-Reyes, Catalina and Cerono, Gabriel and Shi, Yongmei and Rizk-Jackson, Angela and others},
  journal={Bioinformatics},
  pages={btae560},
  year={2024},
  publisher={Oxford University Press}
}

@article{perez2023semantic,
  title={Semantic parsing for conversational question answering over knowledge graphs},
  author={Perez-Beltrachini, Laura and Jain, Parag and Monti, Emilio and Lapata, Mirella},
  journal={arXiv preprint arXiv:2301.12217},
  year={2023}
}

@article{wang2023nlqxform,
  title={NLQxform: A Language Model-based Question to SPARQL Transformer},
  author={Wang, Ruijie and Zhang, Zhiruo and Rossetto, Luca and Ruosch, Florian and Bernstein, Abraham},
  journal={arXiv preprint arXiv:2311.07588},
  year={2023}
}

@article{luo2023chatkbqa,
  title={Chatkbqa: A generate-then-retrieve framework for knowledge base question answering with fine-tuned large language models},
  author={Luo, Haoran and Tang, Zichen and Peng, Shiyao and Guo, Yikai and Zhang, Wentai and Ma, Chenghao and Dong, Guanting and Song, Meina and Lin, Wei and others},
  journal={arXiv preprint arXiv:2310.08975},
  year={2023}
}

@inproceedings{agarwal2024bring,
  title={Bring your own kg: Self-supervised program synthesis for zero-shot kgqa},
  author={Agarwal, Dhruv and Das, Rajarshi and Khosla, Sopan and Gangadharaiah, Rashmi},
  booktitle={Findings of the Association for Computational Linguistics: NAACL 2024},
  pages={896--919},
  year={2024}
}

@article{steinigen2024fact,
  title={Fact Finder--Enhancing Domain Expertise of Large Language Models by Incorporating Knowledge Graphs},
  author={Steinigen, Daniel and Teucher, Roman and Ruland, Timm Heine and Rudat, Max and Flores-Herr, Nicolas and Fischer, Peter and Milosevic, Nikola and Schymura, Christopher and Ziletti, Angelo},
  journal={arXiv preprint arXiv:2408.03010},
  year={2024}
}

@article{jia2024leveraging,
  title={Leveraging Large Language Models for Semantic Query Processing in a Scholarly Knowledge Graph},
  author={Jia, Runsong and Zhang, Bowen and M{\'e}ndez, Sergio J Rodr{\'\i}guez and Omran, Pouya G},
  journal={arXiv preprint arXiv:2405.15374},
  year={2024}
}

@article{morris2023scalable,
  title={The scalable precision medicine open knowledge engine (SPOKE): a massive knowledge graph of biomedical information},
  author={Morris, John H and Soman, Karthik and Akbas, Rabia E and Zhou, Xiaoyuan and Smith, Brett and Meng, Elaine C and Huang, Conrad C and Cerono, Gabriel and Schenk, Gundolf and Rizk-Jackson, Angela and others},
  journal={Bioinformatics},
  volume={39},
  number={2},
  pages={btad080},
  year={2023},
  publisher={Oxford University Press}
}

@misc{liu2023llava,
      title={Improved Baselines with Visual Instruction Tuning}, 
      author={Liu, Haotian and Li, Chunyuan and Li, Yuheng and Lee, Yong Jae},
      publisher={arXiv:2310.03744},
      year={2023},
}

@misc{dolphin_llama3,
author = {{Eric Hartford, Lucas Atkins, and Fernando Fernandes, and Cognitive Computations}},
title  = {{Dolphin Llama 3}},
url    = {https://huggingface.co/dphn/dolphin-2.9-llama3-8b},
note   = {Accessed: 2025-09-11}
}

@article{malartic2024falcon2,
  title={Falcon2-11B Technical Report},
  author={Malartic, Quentin and Chowdhury, Nilabhra Roy and Cojocaru, Ruxandra and Farooq, Mugariya and Campesan, Giulia and Djilali, Yasser Abdelaziz Dahou and Narayan, Sanath and Singh, Ankit and Velikanov, Maksim and Boussaha, Basma El Amel and others},
  journal={arXiv preprint arXiv:2407.14885},
  year={2024}
}

@article{gemma_2024,
    title={Gemma},
    url={https://www.kaggle.com/m/3301},
    DOI={10.34740/KAGGLE/M/3301},
    publisher={Kaggle},
    author={Gemma Team, Thomas Mesnard and Cassidy Hardin and Robert Dadashi and Surya Bhupatiraju and Laurent Sifre and Morgane Rivière and Mihir Sanjay Kale and Juliette Love and Pouya Tafti and Léonard Hussenot and et al.},
    year={2024}
}

@misc{goliath,
author = {{alpindale}},
title  = {{Goliath}},
url    = {https://huggingface.co/alpindale/goliath-120b},
note   = {Accessed: 2025-09-11}
}

@misc{openai2023gpt4,
    title = {GPT-4 Technical Report},
    author = {OpenAI},
    year = {2023},
    howpublished = {\url{https://openai.com/research/gpt-4}},
    note = {Accessed: 2025-09-11}
}

@misc{openai2023gpt4turbo,
    title = {Introducing GPT-4 Turbo},
    author = {OpenAI},
    year = {2023},
    howpublished = {\url{https://openai.com/product/gpt-4}},
    note = {Accessed: 2025-09-11}
}

@article{touvron2023llama,
  title={Llama 2: Open foundation and fine-tuned chat models},
  author={Touvron, Hugo and Martin, Louis and Stone, Kevin and Albert, Peter and Almahairi, Amjad and Babaei, Yasmine and Bashlykov, Nikolay and Batra, Soumya and Bhargava, Prajjwal and Bhosale, Shruti and others},
  journal={arXiv preprint arXiv:2307.09288},
  year={2023}
}

@article{liu2024chatqa,
  title={ChatQA: Surpassing GPT-4 on Conversational QA and RAG},
  author={Liu, Zihan and Ping, Wei and Roy, Rajarshi and Xu, Peng and Lee, Chankyu and Shoeybi, Mohammad and Catanzaro, Bryan},
  journal={arXiv preprint arXiv:2401.10225},
  year={2024}}

@article{dubey2024llama,
  title={The llama 3 herd of models},
  author={Dubey, Abhimanyu and Jauhri, Abhinav and Pandey, Abhinav and Kadian, Abhishek and Al-Dahle, Ahmad and Letman, Aiesha and Mathur, Akhil and Schelten, Alan and Yang, Amy and Fan, Angela and others},
  journal={arXiv preprint arXiv:2407.21783},
  year={2024}
}

@misc{medllama2,
author = {{llSourcell}},
title  = {{medllama2}},
url    = {https://huggingface.co/llSourcell/medllama2_7b},
note   = {Accessed: 2025-09-11}
}

@article{jiang2023mistral,
  title={Mistral 7B},
  author={Jiang, Albert Q and Sablayrolles, Alexandre and Mensch, Arthur and Bamford, Chris and Chaplot, Devendra Singh and Casas, Diego de las and Bressand, Florian and Lengyel, Gianna and Lample, Guillaume and Saulnier, Lucile and others},
  journal={arXiv preprint arXiv:2310.06825},
  year={2023}
}

@article{jiang2024mixtral,
  title={Mixtral of experts},
  author={Jiang, Albert Q and Sablayrolles, Alexandre and Roux, Antoine and Mensch, Arthur and Savary, Blanche and Bamford, Chris and Chaplot, Devendra Singh and Casas, Diego de las and Hanna, Emma Bou and Bressand, Florian and others},
  journal={arXiv preprint arXiv:2401.04088},
  year={2024}
}

@misc{orca-mini,
author = {{pankajmathur}},
title  = {{orca mini 70b}},
url    = {https://huggingface.co/pankajmathur/orca_mini_v3_70b},
note   = {Accessed: 2025-09-11}
}

@article{qwen,
  title={Qwen Technical Report},
  author={Jinze Bai and Shuai Bai and Yunfei Chu and Zeyu Cui and Kai Dang and Xiaodong Deng and Yang Fan and Wenbin Ge and Yu Han and Fei Huang and Binyuan Hui and Luo Ji and Mei Li and Junyang Lin and Runji Lin and Dayiheng Liu and Gao Liu and Chengqiang Lu and Keming Lu and Jianxin Ma and Rui Men and Xingzhang Ren and Xuancheng Ren and Chuanqi Tan and Sinan Tan and Jianhong Tu and Peng Wang and Shijie Wang and Wei Wang and Shengguang Wu and Benfeng Xu and Jin Xu and An Yang and Hao Yang and Jian Yang and Shusheng Yang and Yang Yao and Bowen Yu and Hongyi Yuan and Zheng Yuan and Jianwei Zhang and Xingxuan Zhang and Yichang Zhang and Zhenru Zhang and Chang Zhou and Jingren Zhou and Xiaohuan Zhou and Tianhang Zhu},
  journal={arXiv preprint arXiv:2309.16609},
  year={2023}
}

@misc{starling2023,
    title = {Starling-7B: Improving LLM Helpfulness and Harmlessness with RLAIF},
    author = {Zhu, Banghua and Frick, Evan and Wu, Tianhao and Zhu, Hanlin and Jiao, Jiantao},
    month = {November},
    year = {2023}
}

@misc{vicuna,
author = {{lmsys}},
title  = {{vicuna-33b}},
url    = {https://huggingface.co/lmsys/vicuna-33b-v1.3},
note   = {Accessed: 2025-09-11}
}

@article{xu2023wizardlm,
  title={Wizardlm: Empowering large language models to follow complex instructions},
  author={Xu, Can and Sun, Qingfeng and Zheng, Kai and Geng, Xiubo and Zhao, Pu and Feng, Jiazhan and Tao, Chongyang and Jiang, Daxin},
  journal={arXiv preprint arXiv:2304.12244},
  year={2023}
}

@article{Sequeda2025,
  title = {Knowledge Graphs as a source of trust for LLM-powered enterprise question answering},
  volume = {85},
  ISSN = {1570-8268},
  url = {http://dx.doi.org/10.1016/j.websem.2024.100858},
  DOI = {10.1016/j.websem.2024.100858},
  journal = {Journal of Web Semantics},
  publisher = {Elsevier BV},
  author = {Sequeda,  Juan and Allemang,  Dean and Jacob,  Bryon},
  year = {2025},
  month = may,
  pages = {100858}
}

@article{Wang2024,
  title = {Knowledge Graph Prompting for Multi-Document Question Answering},
  volume = {38},
  ISSN = {2159-5399},
  url = {http://dx.doi.org/10.1609/aaai.v38i17.29889},
  DOI = {10.1609/aaai.v38i17.29889},
  number = {17},
  journal = {Proceedings of the AAAI Conference on Artificial Intelligence},
  publisher = {Association for the Advancement of Artificial Intelligence (AAAI)},
  author = {Wang,  Yu and Lipka,  Nedim and Rossi,  Ryan A. and Siu,  Alexa and Zhang,  Ruiyi and Derr,  Tyler},
  year = {2024},
  month = mar,
  pages = {19206–19214}
}

@article{huang2025survey,
  title={A survey on hallucination in large language models: Principles, taxonomy, challenges, and open questions},
  author={Huang, Lei and Yu, Weijiang and Ma, Weitao and Zhong, Weihong and Feng, Zhangyin and Wang, Haotian and Chen, Qianglong and Peng, Weihua and Feng, Xiaocheng and Qin, Bing and others},
  journal={ACM Transactions on Information Systems},
  volume={43},
  number={2},
  pages={1--55},
  year={2025},
  publisher={ACM New York, NY}
}

\end{document}